\documentclass{article}
\usepackage{fancyhdr}

    \usepackage[preprint]{neurips_2024}

\usepackage[utf8]{inputenc} 
\usepackage[T1]{fontenc}    
\usepackage{hyperref}       
\usepackage{url}            
\usepackage{booktabs}       
\usepackage{amsfonts}       
\usepackage{fancyhdr}       
\newcommand{\footerfont}{\normalfont}
\usepackage{nicefrac}       
\usepackage{microtype}      
\usepackage{xcolor}         
\usepackage{graphicx}
\usepackage{multirow}
\usepackage{CJKutf8}
\definecolor{uclablue}{rgb}{0.15, 0.45, 0.68}
\hypersetup{
    breaklinks,
    citecolor=uclablue,
    colorlinks=true,
}
\usepackage{wrapfig}
\usepackage{float}
\usepackage{subcaption}
\usepackage{placeins}
\usepackage{lipsum} 
\usepackage{tcolorbox}
\usepackage{amsmath}
\usepackage{amssymb}
\usepackage{utfsym}
\usepackage{fontawesome}
\usepackage{xspace}
\usepackage{enumitem}
\usepackage{multirow} 
\tcbuselibrary{breakable}
\usepackage{enumitem}
\usepackage{colortbl}
\usepackage{fancyhdr}

\title{Can  Large Language Models Detect Errors in Long \\ Chain-of-Thought Reasoning?}

\author{
    Yancheng He\textsuperscript{*}$^{1}$, 
    Shilong Li\textsuperscript{*}$^{1}$, 
    Jiaheng Liu\textsuperscript{*,\dag}$^{1}$, 
    Weixun Wang\textsuperscript{*}$^{1}$, 
    Xingyuan Bu$^{1}$, 
    Ge Zhang$^{2}$,\\[10pt]
    \vspace{0.3cm}
    \textbf{Zhongyuan Peng}$^{1}$,
    \textbf{Zhaoxiang Zhang}$^{3}$,
    \textbf{Zhicheng Zheng}$^{1}$, 
    \textbf{Wenbo Su}$^{1}$, 
    \textbf{Bo Zheng}$^{1}$
    \\
    $^{1}$Alibaba Group, $^{2}$M-A-P, $^{3}$CASIA 
    \\[8pt]
    \texttt{\{heyancheng.hyc, ljh411989\}@alibaba-inc.com}
}

\begin{document}

\maketitle
\let\oldthefootnote\thefootnote

\let\thefootnote\relax\footnotetext{*~Equal Contribution. ~~$^\dagger$~Corresponding Author.}
\let\thefootnote\oldthefootnote

\begin{abstract}
Recently, o1-like models have drawn significant attention, where these models produce the long Chain-of-Thought (CoT) reasoning steps to improve the reasoning abilities of existing Large Language Models (LLMs).
In this paper,
to understand the qualities of these long CoTs and measure the critique abilities of existing LLMs on these long CoTs,
we introduce the \textbf{DeltaBench}\footnote{The dataset is available at \url{https://github.com/OpenStellarTeam/DeltaBench}.} including the generated long CoTs from different o1-like models  (e.g., QwQ, DeepSeek-R1) for different reasoning tasks (e.g., Math, Code, General Reasoning), to measure the ability to \textbf{D}etect \textbf{E}rrors in \textbf{L}ong Co\textbf{T} Re\textbf{A}soning.
Based on DeltaBench, we first perform 
fine-grained analysis of the generated long CoTs to discover the effectiveness and efficiency of different o1-like models.
Then, we conduct extensive evaluations of existing process reward models (PRMs) and critic models to detect the errors of each annotated process, which aims to investigate the boundaries and limitations of existing PRMs and critic models.
Finally, we hope that DeltaBench could guide developers to better understand the long CoT reasoning abilities of their models.


\end{abstract}

\section{Introduction}

Large Language Models (LLMs) have witnessed remarkable progress in recent years~\citep{dubey2024llama3, he2024chinese, li2024graphreader,zhang2024mapneo}. Among these advancements, o1-like LLMs have emerged to greatly improve reasoning capabilities by generating long Chain-of-Thought (CoT) reasoning steps~\citep{openai-o1,guo2025deepseek}.


However, despite the growing popularity of o1-like models and their long CoT reasoning approaches, \textit{systematic evaluation of the quality and effectiveness of the generated reasoning chains is not well investigated},
which poses a significant challenge in understanding the capabilities and limitations of these models~\citep{Chen2024DoNT,Yeo2025DemystifyingLC}.
Besides, as LLMs continue to evolve, further improving their performance has become increasingly challenging. One of the key areas of focus in this regard is the development of critique ability,
which measures the capability to provide detailed analysis, constructive suggestions, and refinement feedback suggestions for solutions generated by other models or even themselves~\citep{lan2024criticeval,lin2024criticbench,judgebench,prmbench,Zheng2024ProcessBenchIP}. However, \textit{evaluating the critique abilities of existing LLMs (e.g., critic models or process reward models) on the long CoT reasoning steps has not been explored}.

\begin{figure*}[t]
    \centering
    \includegraphics[width=1.0\textwidth]{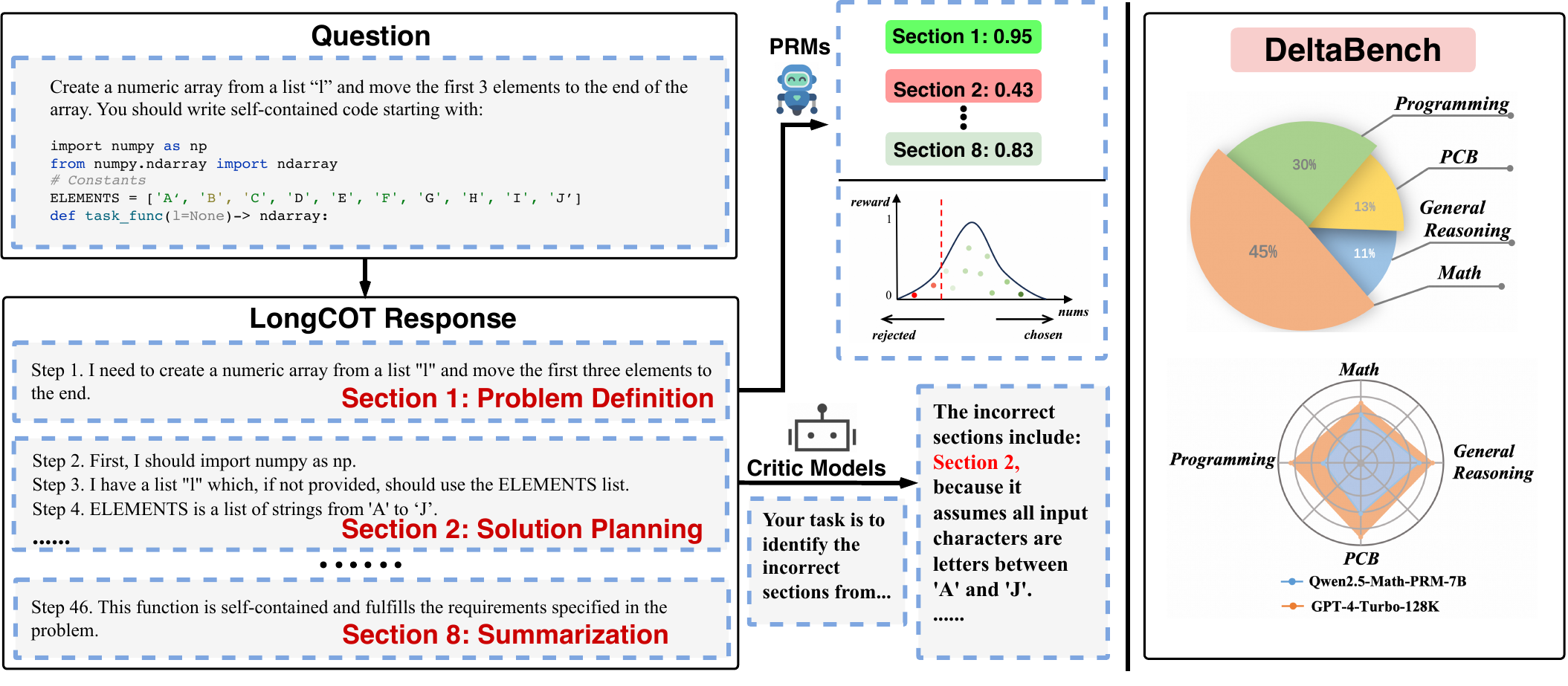}
    \caption{Illustration of the evaluation process for critic models and Process Reward Models (PRMs) for DeltaBench.}
    \label{fig: intro}
\end{figure*}
Therefore, to address the aforementioned challenges, we introduce \textbf{DeltaBench}, 
the first dataset to analyze the qualities of the long CoTs generated by o1-like models and evaluate the critique abilities to \textbf{D}etect \textbf{E}rror in \textbf{L}ong Co\textbf{T} Re\textbf{A}soning of existing critic models and PRMs in Figure \ref{fig: intro}.
Specifically,
in DeltaBench, we first collect a diverse collection of long CoTs generated by various o1-like models (i.e., QwQ, DeepSeek-R1, and Gemini 2.0 Flash Thinking) across different reasoning tasks such as \textbf{Math}, \textbf{Programming}, \textbf{PCB} (physics, chemistry and biology), and \textbf{General Reasoning}.
Then, we divide each long CoT into different sections, where each section denotes an independent subtask. After that, each section is annotated with corresponding labels (e.g., reasoning usefulness, reasoning correctness, and reflection).




Based on DeltaBench,
we first conduct a fine-grained analysis on the efficiency of the generated long CoTs from different o1-like models, and have the following interesting findings:

\begin{itemize}[left=1em]

\item \textbf{Fundamental errors (e.g.,{calculation errors, syntax errors and format errors}) are usually existed in existing o1-like models.}
For example, such errors account for approximately 25\% and 23\% in QwQ-32B-Preview and Gemini 2.0 Flash Thinking,
respectively.

\item \textbf{The proportion of effective reflection is still very low}. Note that the ``effective reflection'' denotes this reflection leads to the right answer.
For example, on average, approximately 67.8\% of the reflections in the collected long CoT responses are useless.

\item \textbf{The long CoT reasoning process is very redundant for existing o1-like models.}
For example, on average, 27\% of the reasoning sections in the collected long CoT response are redundant.



\end{itemize}

After that, we evaluate the critique abilities of LLMs prompted as critic models and PRMs and draw the following insightful observations:

\begin{itemize}[left=1em]

\item \textbf{For existing LLMs and PRMs, the ability to effectively identify errors in long CoT reasoning is {very limited}}. For example, the top-performing model in DeltaBench, GPT-4-turbo-128k, achieves an F1-score of only 40.8\%.

\item \textbf{o1-like models (e.g., o1-mini, o1-preview) do not show any advantage over non-o1-like models on the critique abilities.} For example, the results of o1-preview are lower than the results of GPT-4o-mini.

\item \textbf{The self-critique abilities of existing o1-like models are generally weaker than their critique abilities on other o1-like models.}
For example, DeepSeek-R1 exhibits a 36\% reduction in self-critique performance compared to the critique performance of other o1-like models.

\end{itemize}





\section{DeltaBench}

\begin{figure*}[t]
\centering
\includegraphics[width=1.0\linewidth]{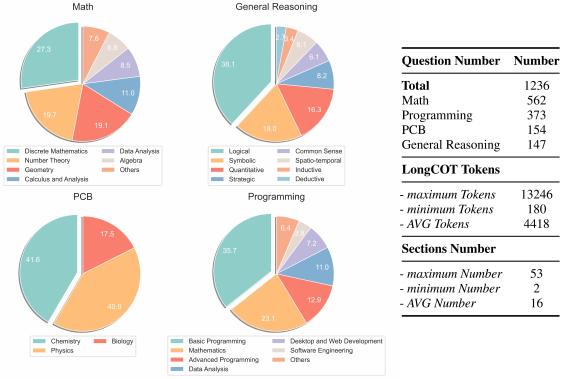}
\caption{Left: Overview of DeltaBench. These pie charts show the distribution of questions in Math, General Reasoning, PCB (Physics, Chemistry and Biology), and Programming. Right:  Statistics of DeltaBench.}
\label{fig: category}
\end{figure*}

In this section, we detail the construction of the DeltaBench dataset, developed to assess a model's capacity to identify and locate errors in long CoT reasoning processes. DeltaBench comprises 1,236 samples across diverse domains, including \textbf{Math}, \textbf{Programming}, \textbf{PCB} (physics, chemistry and biology), and \textbf{General Reasoning}. Each sample encompasses a problem, its corresponding long CoT solution, and comprehensive human annotations. Specifically, the long CoT is divided into sections, and each section includes the following tags: 
\begin{itemize}[left=1em]

\item \textbf{Strategy Shift:} whether this section introduces a new method or strategy attempt. If a new strategy is introduced, the specific step is annotated.

\item \textbf{Reasoning Usefulness:} whether the reasoning in this section is useful. If the process of section can help to lead to the right answer, it considered as useful.

\item \textbf{Reasoning Correctness:} whether this section contains any errors. If an error is present, additional error-related fields are annotated, including the first step number at which the error occurs, explanation and correction.

\item \textbf{Reflection Efficiency:} whether this section contains reflection and whether the reflection is correct. If reflection is present, the step at which the reflection begins is annotated.

\end{itemize}


\subsection{Dataset Construction}

\paragraph{Query collection.}
We extract queries from diverse open-source datasets. Detailed data sources are listed in Appendix \ref{app: data_source}. The domains include math, programming, physics, chemistry, biology, and general reasoning, which comprise 48 subcategories. To ensure the dataset's diversity and balance, we employ a multi-step process:

\begin{itemize}[left=1em]
\item \textbf{Clustering and Deduplication}: Queries are first converted into vector representations using the NV-Embed-v2\footnote{https://huggingface.co/nvidia/NV-Embed-v2} embedding model.  Then, similarity scores are computed between each pair of queries to identify and eliminate duplicates using a predefined threshold. The non-duplicate queries are clustered using DBSCAN~\citep{DBSCN}, resulting in 17,510 unique queries.

\item \textbf{Difficulty Filtering}: For each query, multiple models\footnote{GPT-4o~\citep{gpt4}, Meta-Llama-3-70B-Instruct~\citep{dubey2024llama3}, Meta-Llama-3-8B-Instruct~\citep{dubey2024llama3}, Qwen2.5-72B-Instruct~\citep{qwen2.5}, Qwen2.5-32B-Instruct~\citep{qwen2.5}, and DeepSeek-67B-chat~\citep{deepseek-llm}.} were employed to generate solutions, and difficulty labels were assigned based on the accuracy of the answers produced by these models. Following this, uniform sampling was carried out according to these difficulty labels to ensure a balanced distribution of difficulties.

\item \textbf{Subcategory Sampling}: For each query, GPT-4o is used to classify it into a subcategory. The queries are then uniformly sampled based on these subcategories to ensure diversity.

\end{itemize}

\paragraph{Data Preprocessing.}
We observe low-quality queries exist in open-source datasets. To address this issue, we employed GPT-4 and rule-based filtering strategies to identify and remove these low-quality queries. We have recorded encountered issues. Specific details are shown in Appendix \ref{app: data_preprocess}.

\paragraph{Long CoT Generation.}

We generate long CoT solutions using several open-source o1-like models, such as QwQ-32B-Preview, DeepSeek-R1, and Gemini 2.0 Flash Thinking, with random sampling to enhance diversity. This method ensures a wide range of reasoning processes and captures potential errors models may produce in real-world scenarios, enabling a robust evaluation of error detection capabilities in long CoT reasoning.

\paragraph{Section Split.}

\begin{figure*}[!tbp]
\centering
\includegraphics[width=\linewidth]{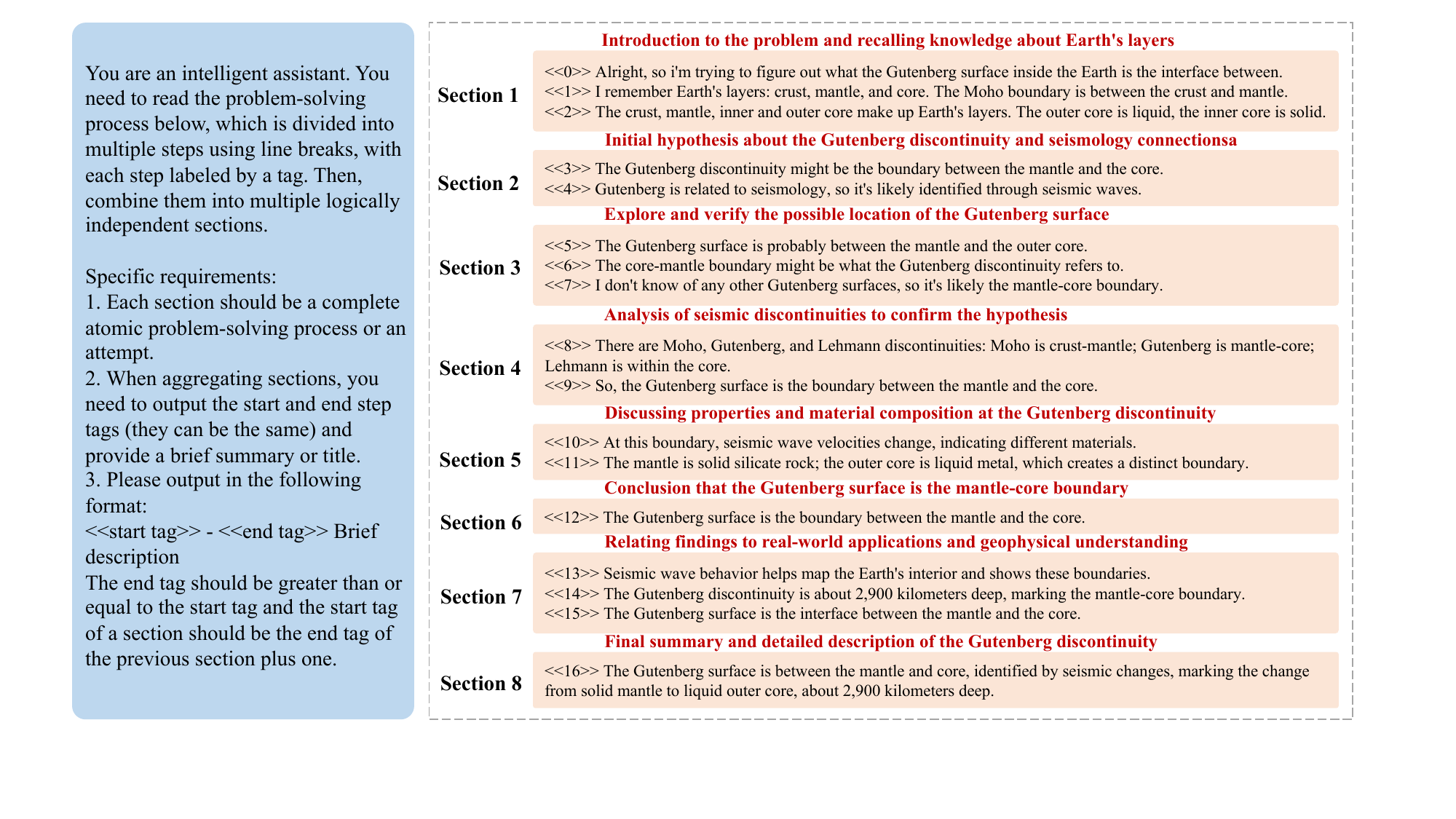}
\caption{An example of section division for long CoT reasoning process.}
\label{fig: section_div}
\end{figure*}


Previous approaches typically divide solutions into steps. However, long CoT responses often contain numerous steps, which significantly increases the difficulty of human annotation, and many of which are either overly granular or lack meaningful contribution to the overall reasoning process. To address this issue, we segment each long CoT response into multiple sections, each representing an independent sub-task, aligning more closely with human cognitive patterns. Specifically, we use the delimiter "\verb|\n\n|" to partition the model’s response into steps first. Then, we employ GPT-4 to identify the start and end steps of each section and generate a brief summary of the content within each section. This approach not only facilitates manual annotation but also enhances the accuracy of the model's segmentation process. The details are provided in Appendix \ref{app: section_div}. 


\subsection{Correctness Assessment}
Before manual annotation, we employ automated methods to assess the correctness of the long CoT results. Domain-specific techniques are used to identify potentially erroneous outputs in Appendix \ref{app:asses} and the evaluation accuracy of each domain and the details are shown in Appendix \ref{app: correct_assess}. 
This process ensures that the data provided for manual annotation are likely to contain errors, which enhances the annotation efficiency.

\subsection{Human Annotation}

\begin{figure}[t]
    \centering
    \includegraphics[width=1.0\textwidth]{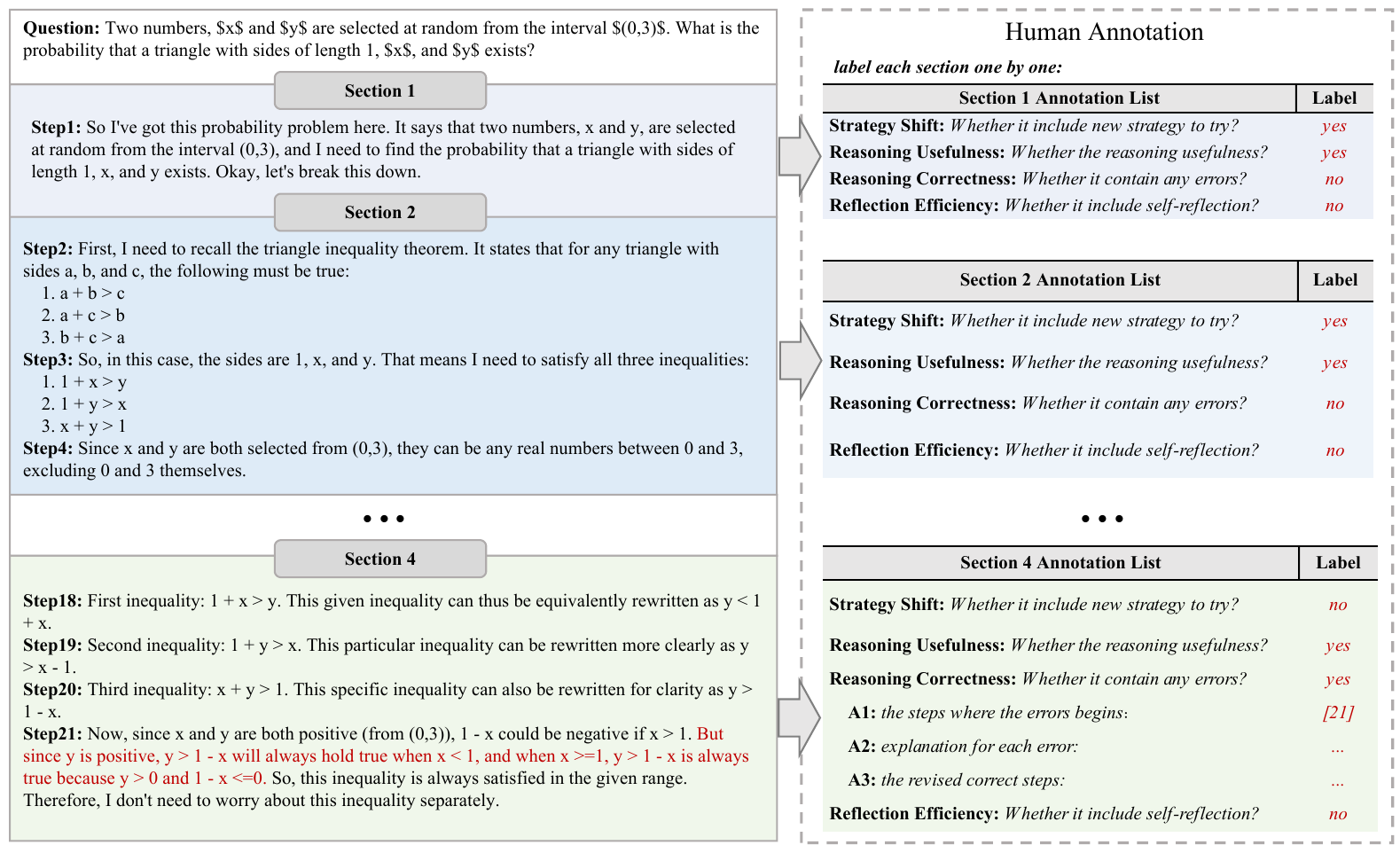}
    \caption{An example of human annotation applied to a mathematical problem-solving process. Annotators are required to annotate each section individually.}
    \label{fig: label_case}
\end{figure}

The data annotation process aims to evaluate the reasoning process of each long CoT response systematically. Each section is assessed for \textbf{strategy shift}, \textbf{reasoning usefulness}, \textbf{reasoning correctness}, and \textbf{reflection efficiency}, as shown in Figure \ref{fig: label_case}. The annotation of whether strategy shift and reflection occurred is to help analyze o1's thinking pattern. The annotation of Reasoning usefulness and error identification is to better analyze and evaluate the performance of system II thinking and further evaluate the critique ability of other models for these problems. 

To ensure high-quality annotations, we recruit Master's and Ph.D. graduates from various disciplines and collaborate with specialized suppliers (See Appendix \ref{app:anno} for more details on the annotation and quality control processes).

\subsection{Dataset Statistics}

\textbf{DeltaBench} contains 1,236 carefully curated samples. These samples are distributed across five major domains and 48 subcategories. The dataset ensures a balance of question types and difficulty levels, incorporating a rich set of long CoT responses. 


\begin{figure}[!htbp]
    \centering
    \begin{subfigure}{0.48\textwidth}
        \centering
        \includegraphics[width=0.9\textwidth]{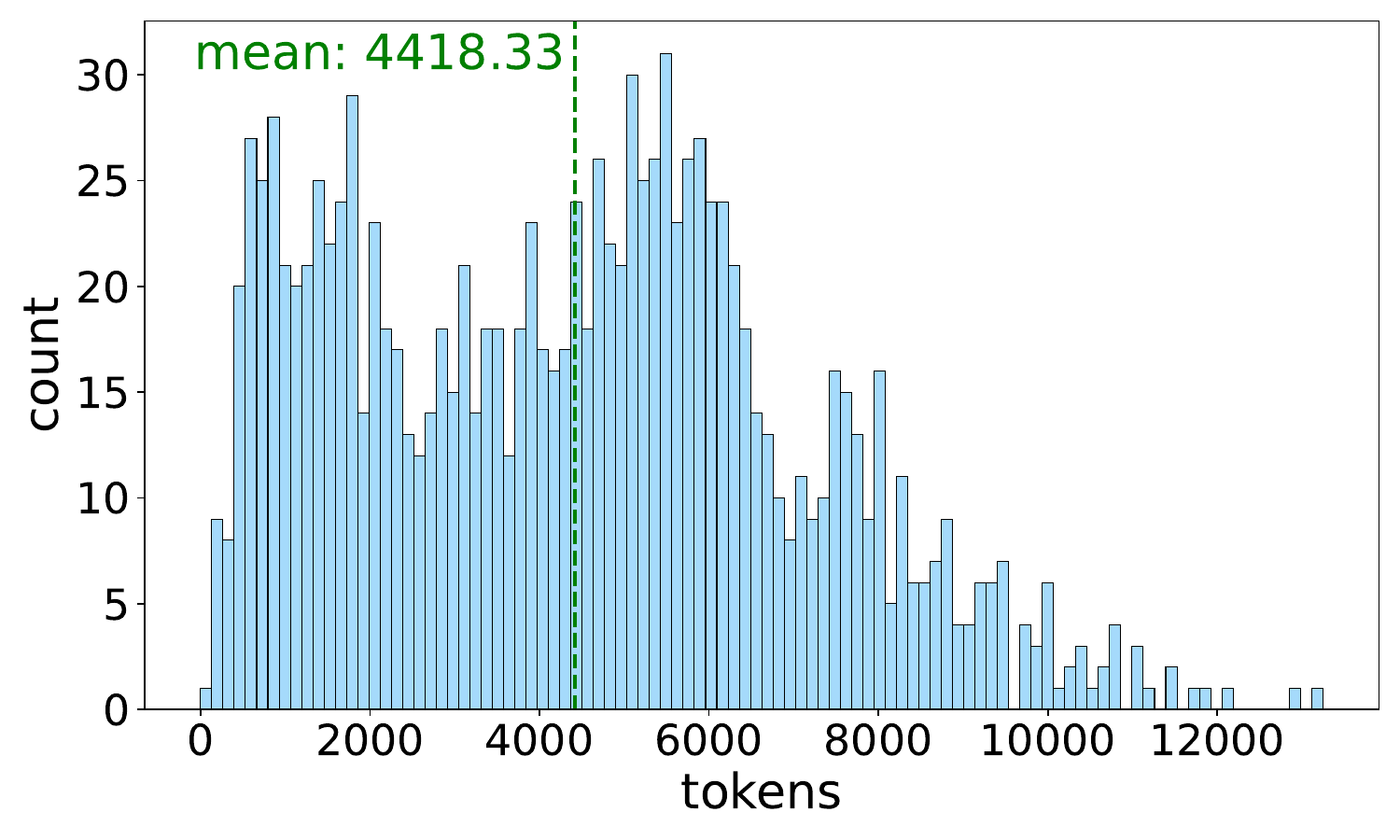}
        \caption{Distribution of long CoT Length.}
        \label{fig: longcot_length_dist}
    \end{subfigure}
    \hfill
    \begin{subfigure}{0.48\textwidth}
        \centering
        \includegraphics[width=0.9\textwidth]{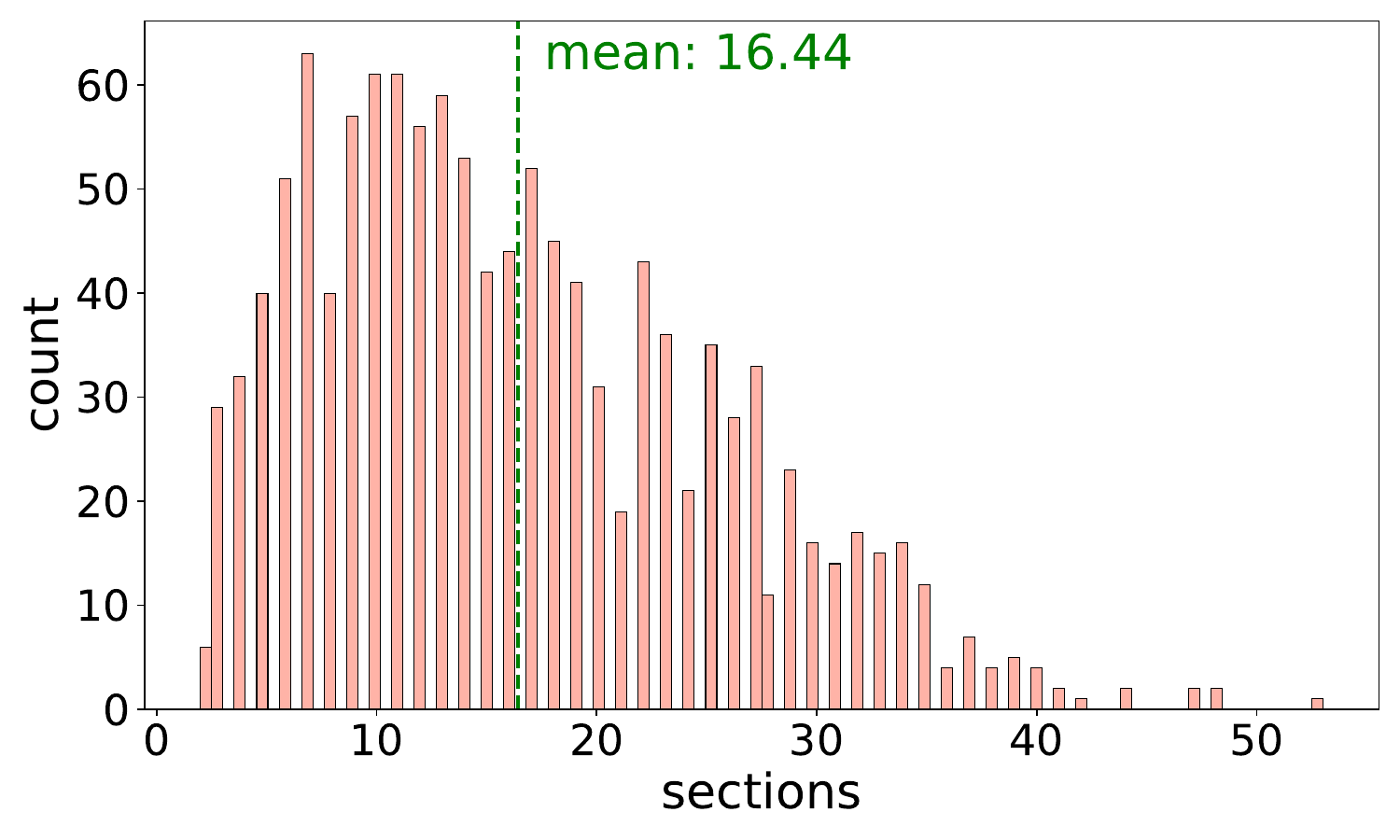}
        \caption{Distribution of the number of long CoT sections.}
        \label{fig: longcot_section_dist}
    \end{subfigure}
    \caption{Distribution of long CoT characteristics.}
    \label{fig: longcot_dist}
\end{figure}

Figure \ref{fig: longcot_dist} shows the distribution of both the length and the number of sections of long CoTs. The distribution is relatively balanced overall, enabling a comprehensive evaluation of the performance of PRMs or critic models across a range of different lengths. Additionally, detailed statistics on the category distribution are provided in Appendix \ref{app: category_distribution}.

\subsection{Evaluation Metrics}
We employ \textbf{recall}, \textbf{precision}, and \textbf{macro-F1 score} for error sections as evaluation metrics. For the PRMs, we utilize an outlier detection technique based on the Z-Score to make predictions. This method was chosen because threshold-based prediction methods determined from other step-level datasets, such as those used in ProcessBench~\citep{Zheng2024ProcessBenchIP}, may not be reliable due to significant differences in dataset distributions, particularly as DeltaBench focuses on long CoT (The details are provided in Appendix \ref{app: other_evaluation}). Outlier detection helps to avoid this bias. The threshold $t$ for determining the correctness of a section is defined as:
$t = \mu - \sigma$,
where $\mu$ is the mean of the rewards distribution across the dataset, and $\sigma$ is the standard deviation. Sections falling below $t$ are predicted as error sections. For critic models, all erroneous sections within a long CoT are prompted to be identified. Given that error sections constitute a smaller proportion than correct sections across the dataset, we use macro-F1 to mitigate the potential impact of the imbalance between positive and negative sections. Macro-F1 independently calculates the F1 score for each sample
and then takes the average, providing a more balanced evaluation metric when dealing with class imbalance.

\subsection{Comparison to Other Benchmarks}

\begin{table}[t]
    \centering
    \resizebox{\textwidth}{!}{
    \begin{tabular}{llcc}
        \toprule
        \textbf{Benchmark} & \textbf{Source} & \textbf{Long CoT} & \textbf{Granularity} \\ 
        \midrule
        JudgeBench&
        \begin{tabular}[c]{@{}l@{}}
            MMLU-Pro~\citep{Wang2024MMLUProAM},
            LiveBench~\citep{White2024LiveBenchAC},\\ 
            LiveCodeBench~\citep{Jain2024LiveCodeBenchHA}
        \end{tabular} & 
        \texttimes & Sample-Level \\ \hline
        CriticBench&
        \begin{tabular}[c]{@{}l@{}}
            GSM8K~\citep{cobbe2021gsm8k},
            CSQA~\citep{Talmor2019CommonsenseQAAQ},\\
            BIGBench~\citep{Srivastava2022BeyondTI}, HumanEval~\citep{Chen2021EvaluatingLL}, etc
        \end{tabular} & 
        \texttimes & Sample-Level \\ \hline
        
        CriticEval & 
        \begin{tabular}[c]{@{}l@{}}
            GSM8K~\citep{cobbe2021gsm8k}, HumanEval~\citep{Chen2021EvaluatingLL}, \\ ChatArena~\citep{Chiang2024ChatbotAA}, etc.
        \end{tabular} & 
        \texttimes & Sample-Level \\\hline
        
        ProcessBench&
        \begin{tabular}[c]{@{}l@{}}
            GSM8K~\citep{cobbe2021gsm8k}, MATH~\citep{Lightman2023LetsVS}, \\
            OlympiadBench~\citep{He2024OlympiadBenchAC},
            Omni-MATH~\citep{Gao2024OmniMATHAU}
        \end{tabular} & 
        \texttimes & Step-Level \\
        
        \midrule
        \textbf{DeltaBench}&
        \begin{tabular}[c]{@{}l@{}}
        AIME, BigCodeBench~\citep{zhuo2024bigcodebench}, 
        KOR-Bench~\citep{Ma2024KORBenchBL}, \\ 
        GPQA~\citep{Rein2023GPQAAG}, etc
        \end{tabular} & 
        \checkmark & Section-Level \\
        \bottomrule
    \end{tabular}

    }
    \vspace{0.3cm}
            \caption{Comparisons between different benchmarks. Sample-level evaluation classifies the entire model response as correct or incorrect. Step-level evaluation assesses the correctness of individual reasoning steps. Section-level evaluation evaluates the correctness of reasoning sections which is a more appropriate granularity for long CoT response.}
                \label{table:benchmark_compare}

\end{table}
In Table \ref{table:benchmark_compare}, 
DeltaBench has the following features:
(1) We focus on difficult questions, providing a more challenging critical evaluation; (2) We utilize long CoT responses, enabling the assessment of a model's ability to identify errors within complex reasoning processes; (3) We evaluate the model's capability to identify all errors in the reasoning process, rather than just the first error or a binary classification of correctness on sample level,
which can provide a fine-grained analysis of long CoTs.

\section{Analysis}

\subsection{Error Analysis of o1-like Models}

\paragraph{Error Type Lists}
We classify the errors that occur during the system II thinking process into 8 major aspects and 23 specific error types based on the manual annotations, including understanding errors, reasoning errors, reflection errors, summary errors, etc. For detailed information about the error categories, see Appendix \ref{app: error_classification}.

\paragraph{What Are the Most Common Errors Across Domains?}

\begin{figure}[t]
    \centering
    \resizebox{1.0\textwidth}{!}
    {\includegraphics{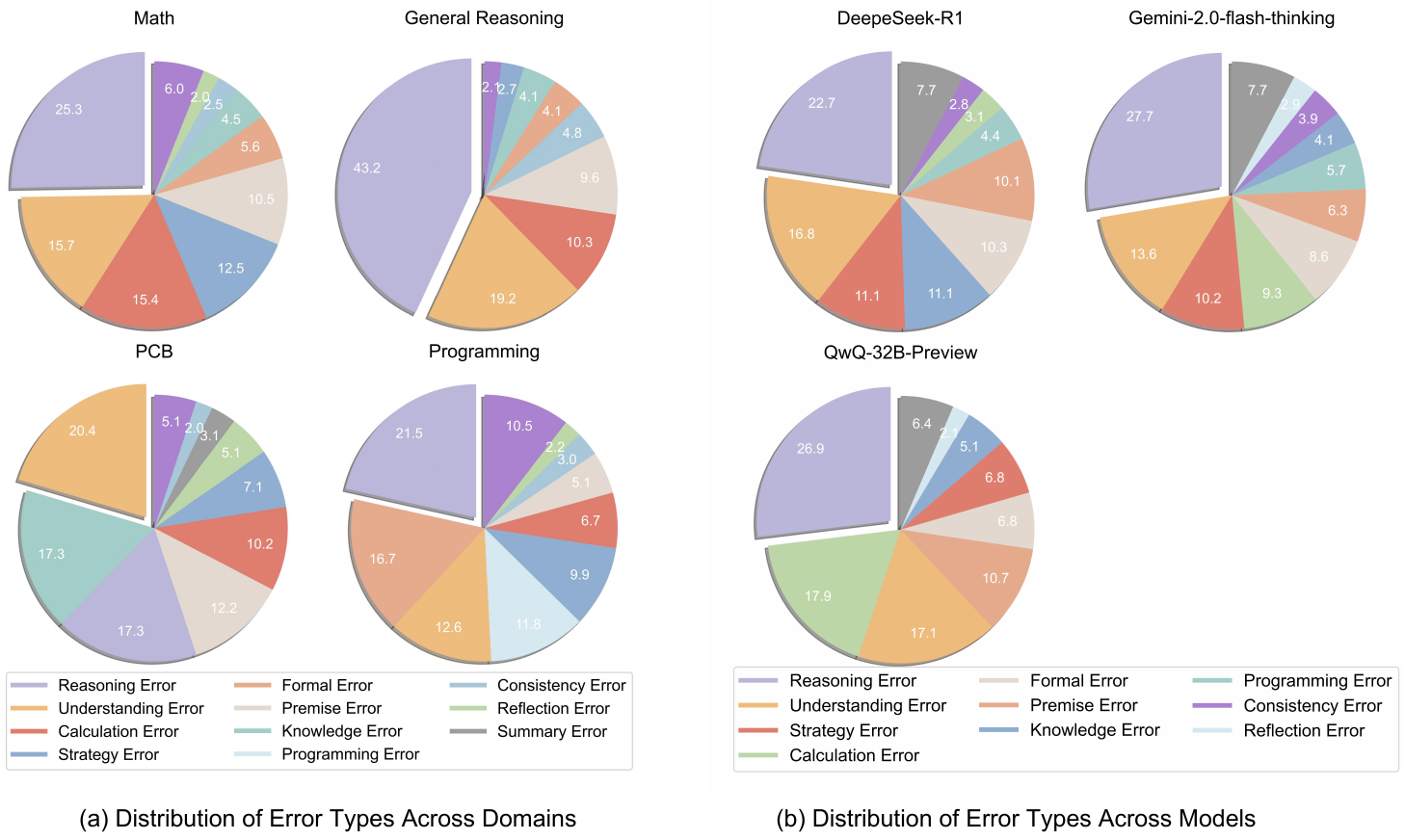}}
    \caption{Distribution of error types across different domains and models.}
    \label{fig: error_type}
\end{figure}

To analyze the characteristics of error distribution in different domains, we performed a uniform sampling of the data based on the model, the domain, and the query difficulty. Figure \ref{fig: error_type} shows the error distribution across different domains, here are some key findings:

\begin{itemize}[left=1em]
\item \textbf{Math:} The most frequent error type is \textit{Reasoning Error}(25.3\%), followed by \textit{Understanding Error}(15.7\%) and \textit{Calculation Error}(15.4\%). This indicates that while the models often struggle with logical reasoning and problem understanding, low-level computational mistakes also remain a significant issue.

\item \textbf{Programming}: 
\textit{Reasoning Error} (21.5\%) is the most common, followed by \textit{Formal Error} (16.7\%) and \textit{Understanding Error} (12.6\%). The high frequency of \textit{Formal Error} and \textit{Programming Error} (11.8\%) underscores the models' struggles with code-specific details and implementation. 

\item \textbf{PCB}: 
The dominant error types are \textit{Understanding Error} (20.4\%) and \textit{Knowledge Error} (17.3\%), closely followed by \textit{Reasoning Error} (17.3\%). This suggests that the main challenge for current models in the fields of physics, chemistry and biology is to understand field-specific concepts and accurately apply relevant knowledge.

\item \textbf{General Reasoning}: \textit{Reasoning Error} is the most prevalent, accounting for 43\%, followed by comprehension errors, accounting for 19\%, showing that logical reasoning is the primary bottleneck.

\end{itemize}

\paragraph{What Are the Model-Specific Error Patterns?}


We also analyzed errors specific to individual models, providing further insights into model weaknesses, as illustrated in Figure \ref{fig: error_type_model}. The error distributions reveal distinct patterns for each model, highlighting their unique strengths and areas for improvement. Here are some key findings:

\begin{itemize}[leftmargin=4mm]

\item \textbf{DeepSeek-R1} exhibits its most pronounced weakness in \textit{Reasoning Errors} (22.7\%), indicating challenges in constructing coherent and accurate logical reasoning paths. However, it demonstrates relative strength in handling fundamental tasks, with minimal \textit{Calculation Errors} (3.1\%) and \textit{Programming Errors} (4.4\%).


\item \textbf{QwQ-32B-Preview} excels at identifying correct problem-solving approaches. However, its effectiveness is significantly hindered by deficiencies in handling finer details, particularly in \textit{Calculation Errors} (17.9\%)



\end{itemize}

\begin{tcolorbox}[colback=white!95!gray, colframe=gray!70!black,  title=Key Finding for Error Type]
The primary bottleneck of current models remains reasoning ability. However, detailed errors like calculation and formal mistakes also contribute significantly.
\end{tcolorbox}

\subsection{Reflection Analysis of o1-like Models}

\begin{figure}[t]
    \centering
    \includegraphics[width=0.95\textwidth]{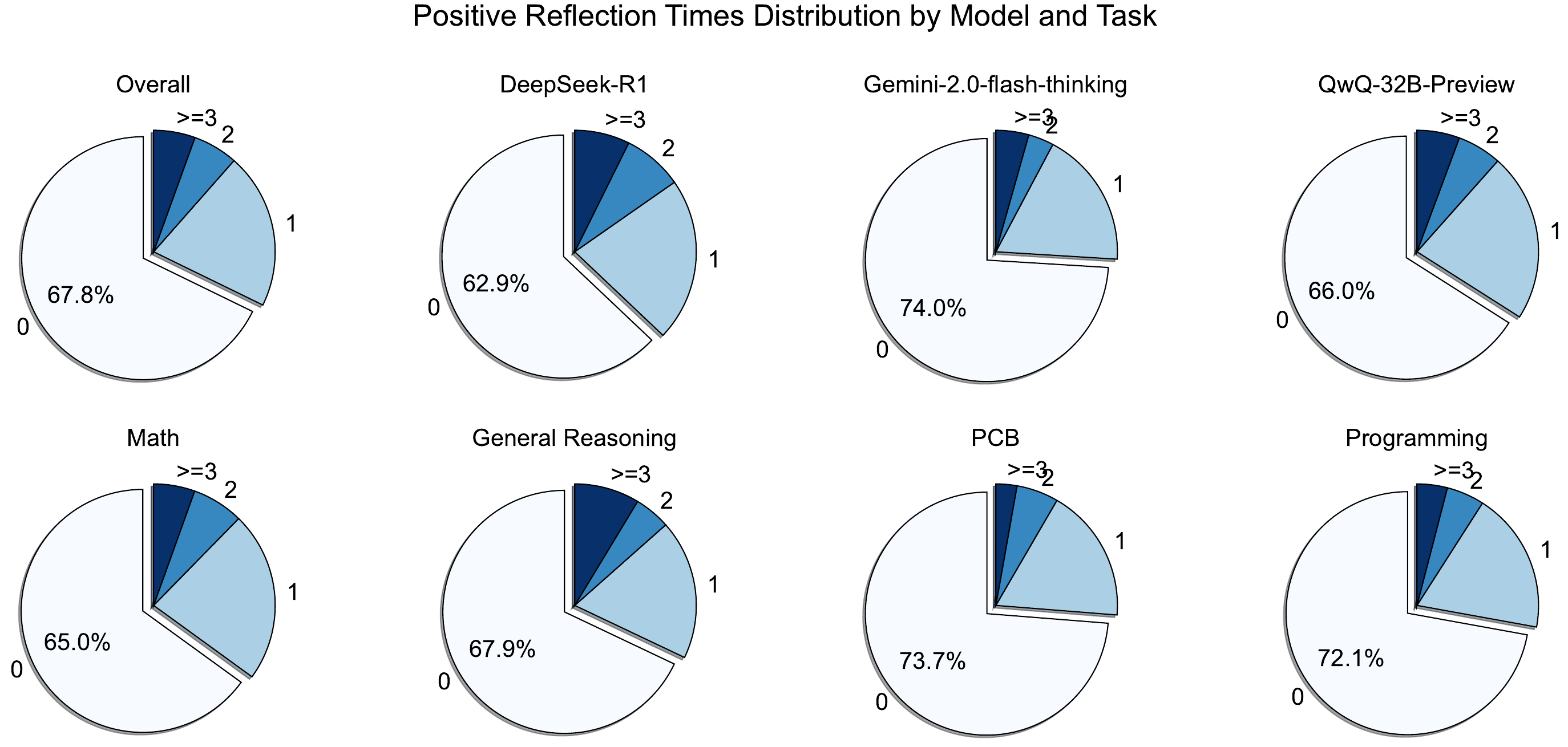}
    \caption{Distribution of effective reflection times by models and domains on a sample level. The segments within each pie chart represent how many times effective reflection occurs in one sample, with segment `0' indicating there is no effective reflection.}
    \label{fig: error_type_model}
\end{figure}

\paragraph{Statistics.}
We also conduct a analysis of the total number of reflections and the proportion of effective reflections in the long CoT output of all questions (including questions answered correctly and incorrectly by the model). 

\paragraph{How Effective Are Model Reflections Across Different Models and Domains?}
We classify samples with reflections based on the number of valid reflections to evaluate the ability to produce valid reflections. Specifically, we label samples as \texttt{0} if no valid reflections occur, and \texttt{1}, \texttt{2}, or \texttt{>=3} for samples with one, two, or three and more valid reflections, respectively(all statistical analyses were performed under strictly controlled conditions, ensuring uniform sampling and balanced tasks for a fair comparison). In Figure \ref{fig: error_type_model}, {DeepSeek-R1} exhibits the highest proportion of effective reflections, and the models show a notably higher rate of effective reflections in the {math} domain. However, the overall proportion of valid reflections across all models remains relatively low, ranging between 30\% and 40\%. This suggests that the reflection capabilities of current models require further improvement.

\begin{tcolorbox}[colback=white!95!gray, colframe=gray!70!black,  title=Key Finding for Reflection]
Despite frequent reflection attempts, the proportion of effective reflections remains low across models, and  DeepSeek-R1 achieves the highest rate of valid reflections.
\end{tcolorbox}

\subsection{Effective Reasoning of o1-like Models}

\begin{figure}[t]
    \centering
    \includegraphics[width=0.98\textwidth]{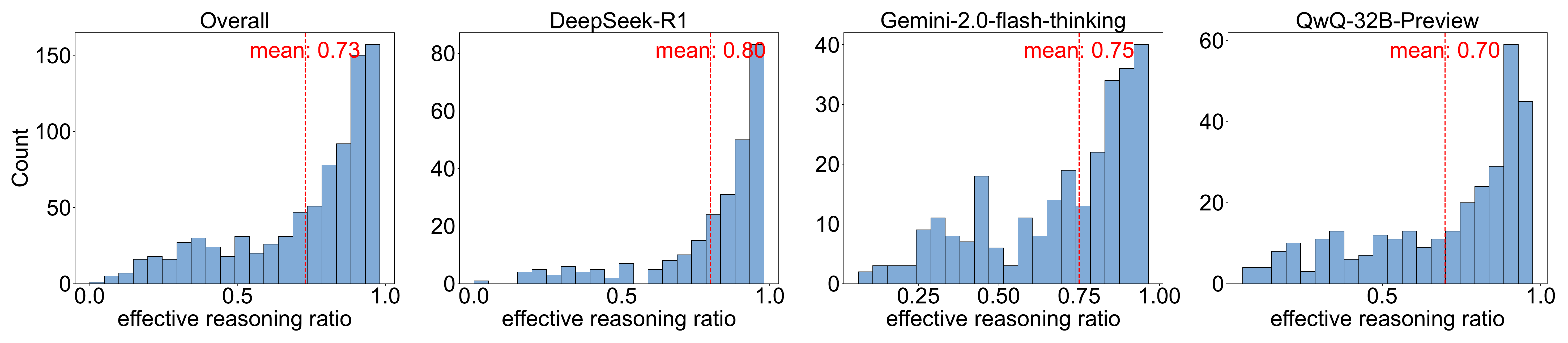}
    \caption{Distribution of effective reasoning ratios.}
    
    \label{fig: effetive_reasoning}
\end{figure}

\paragraph{Statistics.} 
Human annotators evaluate the usefulness of the reasoning in each section, enabling us to calculate the proportion of valid reasoning in each response. As illustrated in Figure \ref{fig: effetive_reasoning}, each graph shows the distribution of effective reasoning ratios for a particular model. The red dashed line in each graph indicates the average effective reasoning ratio.

\paragraph{What Proportion of Reasoning in Long CoT Responses is Effective?}
On average, only 73\% of the reasoning in the collected long CoT responses is useful, highlighting significant redundancy issues. Among the models analyzed, \textit{QwQ-32B-Preview} exhibited the lowest proportion of effective reasoning at 70\%, while \textit{DeepSeek-R1} achieved a notably higher proportion compared to the others, demonstrating superior reasoning efficiency.

\begin{tcolorbox}[colback=white!95!gray, colframe=gray!70!black,  title=Key Finding for Reasoning Efficiency]
On average, 27\% of reasoning in long CoT responses we collected is redundant, and DeepSeek-R1 outperforms others in reasoning efficiency.
\end{tcolorbox}
\vspace{-3mm}

\subsection{Reasoning Process Analysis}

\begin{figure}[t]
    \centering
    \includegraphics[width=0.8\textwidth]{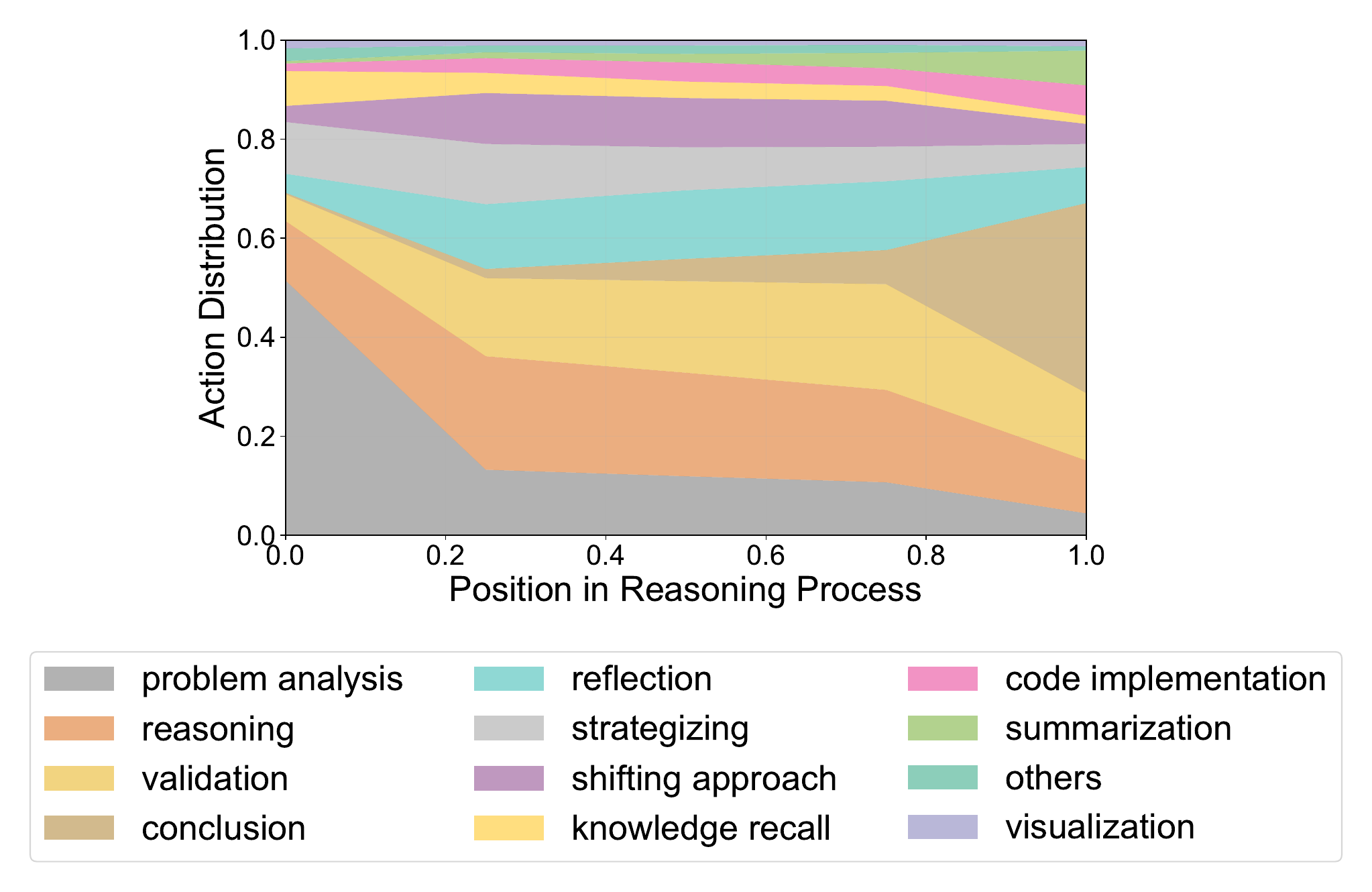}
    \caption{Distribution of different task types throughout the progress of a long CoT response.}
    \vspace{-3mm}
    
    \label{fig: action_roles}
\end{figure}

Figure ~\ref{fig: action_roles} shows the distribution of each section's action roles in the system II thinking process of the o1-like models. Initially, problem analysis dominates, indicating that the model initially focuses on understanding the requirements and constraints of the problem. As the solution progresses, cognitive activities diversify significantly, with reflection and validation becoming more prominent. In the later part of the reasoning, the distribution of conclusion and summarization gradually increases. 

\subsection{Results on DeltaBench}

\begin{table*}[!t]
\centering
\resizebox{1.0\textwidth}{!}{%
    \begin{tabular}{cccccccccccccccc}
    \toprule
    \multirow{2}{*}{\textbf{Model}} & \multicolumn{3}{c}{\textbf{Overall}} & \textbf{Math} & \textbf{Code} & \textbf{PCB} & \textbf{General} \\
    \cmidrule(lr){2-4} \cmidrule(lr){5-5} \cmidrule(lr){6-6} \cmidrule(lr){7-7} \cmidrule(lr){8-8}
     & \textbf{\textit{Recall}} & \textbf{\textit{Precision}} & \textbf{\textit{F1}} & \textbf{\textit{F1}} & \textbf{\textit{F1}} & \textbf{\textit{F1}} & \textbf{\textit{F1}} \\
    \midrule
    \multicolumn{8}{c}{\textbf{\textit{Process Reward Models (PRMs)}}} \\
    \midrule
    \rowcolor[rgb]{ .988,  .949,  .8} Qwen2.5-Math-PRM-7B & \textbf{30.30} & \textbf{34.96} & \textbf{29.22}  &  \textbf{29.64} & \textbf{23.76} & \underline{31.09} & \underline{34.19}   \\
    \rowcolor[rgb]{ .988,  .949,  .8} Qwen2.5-Math-PRM-72B & \underline{28.16} & \underline{29.37} & \underline{26.38}  & \underline{24.16} & \underline{22.02} & \textbf{31.14} & \textbf{35.83}  \\
    \rowcolor[rgb]{ .988,  .949,  .8} Llama3.1-8B-PRM-Deepseek-Data & 11.7 & 15.59 & 12.02 &  12.28 & 10.95 & 16.76 & 12.59  \\
    \rowcolor[rgb]{ .988,  .949,  .8} Llama3.1-8B-PRM-Mistral-Data & 9.64 & 11.21 & 9.45 & 9.40 & 10.72 & 13.43 & 12.40  \\
    \rowcolor[rgb]{ .988,  .949,  .8} Skywork-o1-Qwen-2.5-1.5B & 3.32 & 3.84 & 3.07 & 1.30 & 6.66 & 5.43 & 7.87  \\
    \rowcolor[rgb]{ .988,  .949,  .8} Skywork-o1-Qwen-2.5-7B & 2.49 & 2.22 & 2.17 & 0.78 & 6.28 & 6.02 & 3.11  \\
    \midrule
     \multicolumn{8}{c}{\textbf{\textit{LLM as Critic Models}}} \\
    \midrule
    \rowcolor[rgb]{ .922,  .89,  .988} GPT-4-turbo-128k & \textbf{57.19} & \textbf{37.35} & \textbf{40.76} & \textbf{37.56} & \textbf{43.06} & \underline{45.54} & \underline{42.17} \\
    \rowcolor[rgb]{ .922,  .89,  .988} GPT-4o-mini & \underline{49.88} & 35.37 & \underline{37.82} & \underline{33.26} & 37.95 & \textbf{45.98} & \textbf{46.39} \\
    \rowcolor[rgb]{ .922,  .89,  .988} Doubao-1.5-Pro & 39.68 & \underline{37.02} & 35.25 & 32.46 & \underline{39.47} & 33.53 & 37.00 \\
    \rowcolor[rgb]{ .922,  .89,  .988} GPT-4o & 36.52 & 32.48 & 30.85 & 28.61 & 28.53 & 39.25 & 36.50 \\
    \rowcolor[rgb]{ .922,  .89,  .988} Qwen2.5-Max & 36.11 & 30.82 & 30.49 & 26.73 & 32.81 & 39.49 & 29.54 \\
    \rowcolor[rgb]{ .922,  .89,  .988} Gemini-1.5-pro & 35.51 & 30.32 & 29.59 & 26.56 & 28.20 & 40.13 & 33.66 \\
    \rowcolor[rgb]{ .922,  .89,  .988} DeepSeek-V3 & 32.33 & 28.13 & 27.33 & 27.04 & 27.73 & 27.35 & 27.45 \\
    \rowcolor[rgb]{ .922,  .89,  .988} Llama-3.1-70B-Instruct & 32.22 & 28.85 & 27.67 & 21.49 & 32.13 & 28.45 & 39.18 \\
    \rowcolor[rgb]{ .922,  .89,  .988} Qwen2.5-32B-Instruct & 30.12 & 28.63 & 26.73 & 22.34 & 31.37 & 33.78 & 24.37 \\
    \rowcolor[rgb]{ .882,  .949,  .89} DeepSeek-R1 & 29.20 & 32.66 & 28.43 & 24.17 & 29.28 & 34.78 & 35.87 \\
    \rowcolor[rgb]{ .882,  .949,  .89} o1-preview & 27.92 & 30.59 & 26.97 & 22.19 & 28.09 & 33.11 & 35.94 \\
    \rowcolor[rgb]{ .922,  .89,  .988} Qwen2.5-14B-Instruct & 26.64 & 27.27 & 24.73 & 21.51 & 29.05 & 29.98 & 20.59 \\
    \rowcolor[rgb]{ .922,  .89,  .988} Llama-3.1-8B-Instruct & 25.71 & 28.01 & 24.91 & 18.12 & 32.17 & 27.30 & 29.93 \\
    \rowcolor[rgb]{ .882,  .949,  .89} o1-mini & 22.90 & 22.90 & 19.89 & 16.71 & 21.70 & 20.37 & 26.94 \\
    \rowcolor[rgb]{ .922,  .89,  .988} Qwen2.5-7B-Instruct & 21.99 & 19.61 & 18.63 & 11.61 & 25.92 & 29.85 & 15.18 \\
    \rowcolor[rgb]{ .882,  .949,  .89} DeepSeek-R1-Distill-Qwen-32B & 17.19 & 18.65 & 16.28 & 13.02 & 23.55 & 15.05 & 11.56 \\
    \rowcolor[rgb]{ .882,  .949,  .89} DeepSeek-R1-Distill-Qwen-14B & 12.81 & 14.54 & 12.55 & 9.40 & 18.36 & 10.44 & 12.01 \\
    \bottomrule
    \end{tabular}
}
\caption{Experimental results of PRMs and critic models on DeltaBench. \textbf{Bold} indicates the best results within the same group of models, while \underline{ underline} indicates the second best.}
\label{tab: main}
\end{table*}


\noindent\textbf{Baseline Models.}
For the \textbf{PRMs}, we select the following models: Qwen2.5-Math-PRM-7B\footnote{\href{https://huggingface.co/Qwen/Qwen2.5-Math-PRM-7B}{Qwen/Qwen2.5-Math-PRM-7B}}, Qwen2.5-Math-PRM-72B\footnote{\href{https://huggingface.co/Qwen/Qwen2.5-Math-PRM-72B}{Qwen/Qwen2.5-Math-PRM-72B}}, Llama3.1-8B-PRM-Deepseek-Data\footnote{\href{https://huggingface.co/RLHFlow/Llama3.1-8B-PRM-Deepseek-Data}{RLHFlow/Llama3.1-8B-PRM-Deepseek-Data}}, Llama3.1 -8B-PRM-Mistral-Data\footnote{\href{https://huggingface.co/RLHFlow/Llama3.1-8B-PRM-Mistral-Data}{RLHFlow/Llama3.1-8B-PRM-Mistral-Data}}, Skywork-o1-Open-PRM- Qwen-2.5-1.5B\footnote{\href{https://huggingface.co/Skywork/Skywork-o1-Open-PRM-Qwen-2.5-1.5B}{Skywork/Skywork-o1-Open-PRM-Qwen-2.5-1.5B}}, and Skywork-o1-Open-PRM-Qwen-2.5-7B\footnote{\href{https://huggingface.co/Skywork/Skywork-o1-Open-PRM-Qwen-2.5-7B}{Skywork/Skywork-o1-Open-PRM-Qwen-2.5-7B}}. 
We select a group of the most advanced open-source and closed-source LLMs to serve as \textbf{critic models} for evaluation, which includes various GPT-4~\citep{gpt4} variants (such as GPT-4-turbo-128K, GPT-4o-mini, GPT-4o), the Gemini model~\citep{Reid2024Gemini1U}(Gemini-1.5-pro), several Qwen models~\citep{qwen2.5} (such as Qwen2.5-32B-Instruct and Qwen2.5-14B-Instruct), Doubao-1.5-Pro~\citep{doubao2025}
and o1 models~\citep{openai-o1} (o1-preview-0912, o1-mini-0912).

\subsubsection{Main Results}
In Table \ref{tab: main},
we provide the results of different LLMs on DeltaBench. 
For PRMs, we have the following observations: (1). Existing PRMs usually achieve low performance, which indicates that existing PRMs cannot identify the errors in long CoTs effectively and it is necessary to improve the performance of PRMs. (2). Larger PRMs
do not lead to better performance. For example, the Qwen2.5-Math-PRM-72B is inferior to wen2.5-Math-PRM-7B.
For critic models, we have the following findings: (1)
GPT-4-turbo-128k archives the best critique results, which is better than other models (e.g., GPT-4o) a lot in DeltaBench. (2) For o1-like models (e.g., DeepSeek-R1, o1-mini, o1-preview), we observe that the results of these models are not superior to non-o1-like models, with the performance of o1-preview is even lower than Qwen2.5-32B-Instruct.
A detailed analysis of underperforming models is provided in Appendix \ref{app: underperforming}.


\subsubsection{Further Analysis}

\paragraph{Effect of Long CoT Length.}
\begin{figure}[t]
    \centering
    \includegraphics[width=1.0\textwidth]{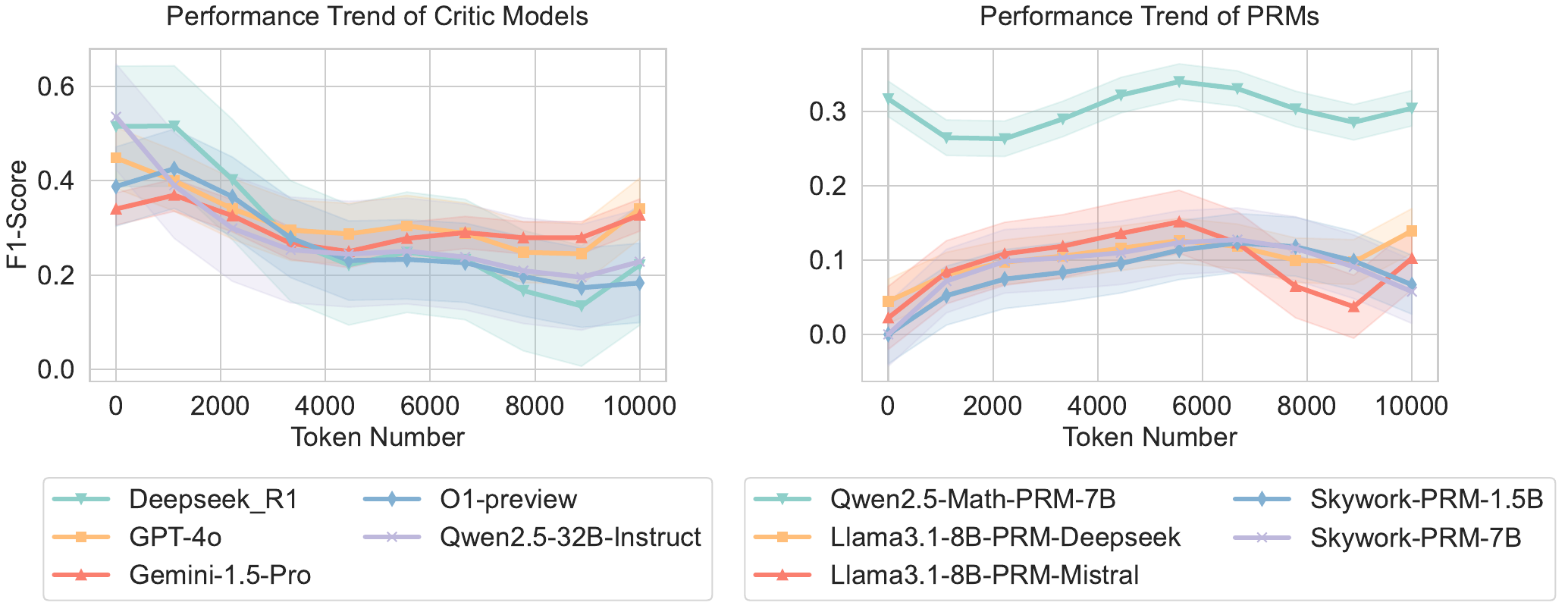}
    \caption{The effect of long CoT length.}
    \label{fig: crtic1}
\end{figure}
In Figure \ref{fig: crtic1}, we compare the average F1-Score performance of critic models and PRMs across varying LongCoT token lengths. 
For critic models, the performance notably declines as token length increases. Initially, models like Deepseek-R1 and GPT-4o exhibit strong performance with shorter sequences (1-3k tokens). However, as token length increases to mid-ranges (4-7k tokens), there is a marked decrease in performance across all models. This trend highlights the growing difficulty for critic models to maintain precision and recall as long CoT response become longer and more complex, likely due to the challenge of evaluating lengthy model outputs. In contrast, PRMs demonstrate greater stability across token lengths, as they evaluate sections sequentially rather than processing the entire output at once. Despite this advantage, PRMs achieve lower overall scores compared to critic models on our evaluation set.

\begin{tcolorbox}[colback=white!95!gray, colframe=gray!70!black, title=Key Finding]
  Critic models exhibit significant performance degradation with longer contexts, while PRMs demonstrate consistent evaluation capability across varying lengths.
\end{tcolorbox}

\paragraph{Performance Analysis Across Different Error Types.}
\begin{figure}[t]
    \centering
    \includegraphics[width=0.9\textwidth]{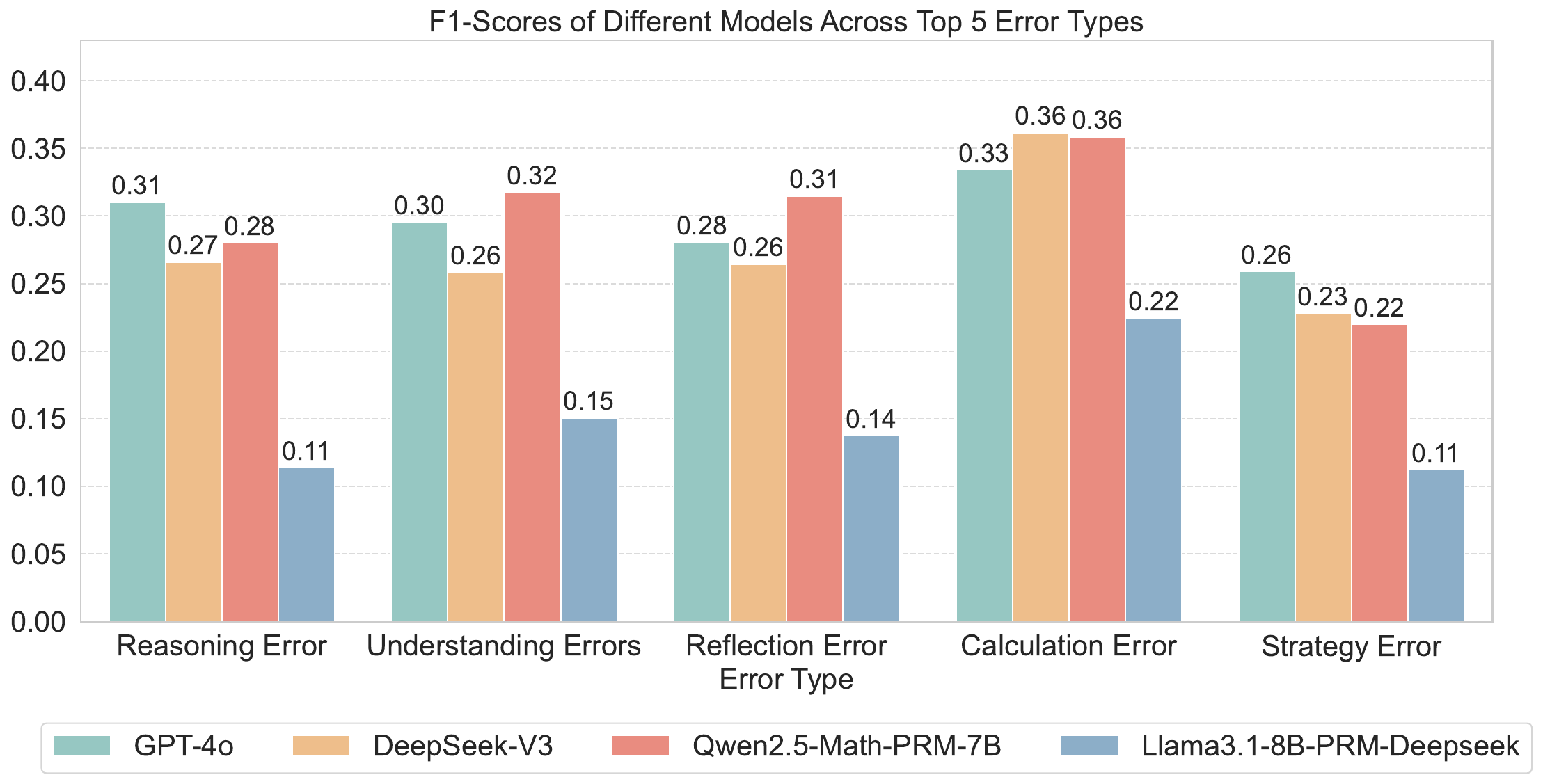}
    \caption{Results of different LLMs on top-5 errors.}
    \label{fig: top_models_per_task}
\end{figure}
Figure \ref{fig: top_models_per_task} shows the performance of different models on the five most common error types. In terms of error types, most models demonstrate the highest accuracy in recognizing calculation errors. Conversely, the recognition of strategy errors is generally the weakest. In terms of models, there is significant variation in the ability of individual models to recognize different error types. For instance, DeepSeek-V3 achieves an F1 of 36\% on calculation errors but only 23\% on strategy errors. Meanwhile, Llama3.1-8B-PRM-Deepseek performs poorly, with an F1 score of 22\% on calculation errors, and shows a significant decline in performance across the other four error types. This highlights the limited generalization capabilities of most models when recognizing various error types.

\begin{tcolorbox}[colback=white!95!gray, colframe=gray!70!black, title=Key Finding]
  Models exhibit strong performance on calculation errors but struggle with strategy errors, revealing limited generalization across error types.
\end{tcolorbox}

\begin{table}[!ht]
    \centering
    \begin{tabular}{cccc}
    \toprule
        \multirow{2}{*}{Model} & \multicolumn{3}{c}{HitRate@$k$ - Avg(\%)} \\ \cline{2-4}
                           & $k=1$ & $k=3$ & $k=5$ \\ 
                           \midrule
        Qwen2.5-Math-PRM-7B & \textbf{49.15} & \textbf{69.14} & \textbf{83.14} \\
        Qwen2.5-Math-PRM-72B & \underline{41.13} & \underline{62.70} & \underline{75.73} \\ 
        Llama3.1-8B-PRM-Deepseek-Data & 12.63 & 48.62 & 69.78 \\
        Llama3.1-8B-PRM-Mistral-Data & 8.99 & 42.97 & 65.33 \\
        Skywork-o1-Open-PRM-Qwen-2.5-1.5B & 31.90 & 53.82 & 69.23 \\
        Skywork-o1-Open-PRM-Qwen-2.5-7B & 31.58 & 52.59 & 69.16 \\
        \bottomrule
    \end{tabular}
    \vspace{+3mm}
    \caption{Results of HitRate@$k$. Bold and underlined results indicate the best and the second best.}
\label{tab: hitrate}
\end{table}

\paragraph{Analysis on HitRate evaluation for PRMs.}

\begin{figure}[t]
    \centering
    \includegraphics[width=\textwidth]{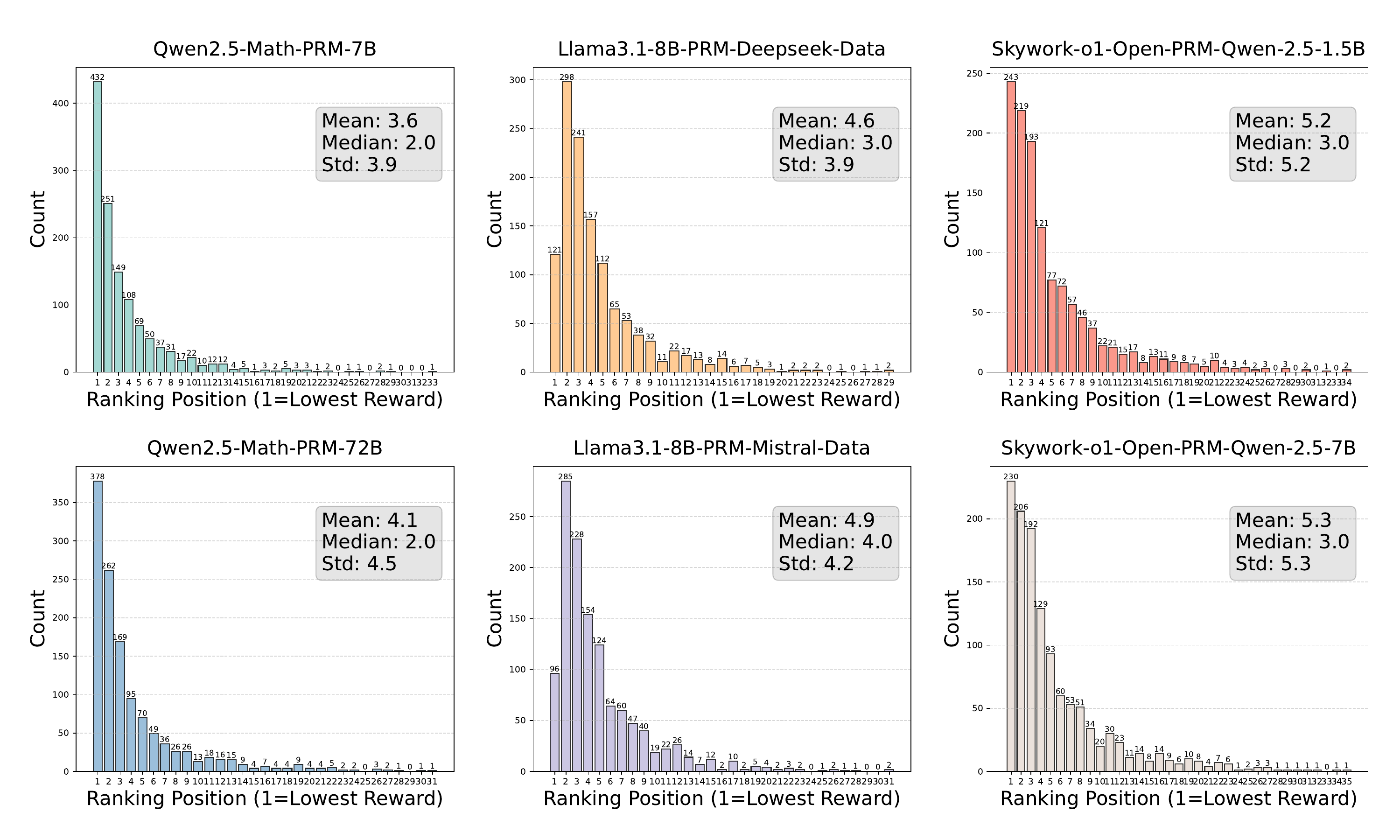}
    \caption{Ranking of rewards for the first incorrect section for different PRMs.}
    \label{fig: prm_rank}
\end{figure}

To better measure the ability of PRMs to identify erroneous sections in long CoTs, we use HitRate@$k$ to evaluate PRMs. Specifically, within a sample, we rank the sections in ascending order based on the rewards given by the PRM, select the smallest $k$ sections, and calculate the recall rate for the erroneous sections among them. Specifically, we define the sorted sections as $S = \{s_1, s_2, \ldots, s_n\}$, with $E$ being the set of erroneous sections. We select the top $k$ sections, denoted as $S_k = \{s_1, s_2, \ldots, s_k\}$. The HitRate@$k$ is  calculated as:
\begin{align}
\text{HitRate@}k = \frac{|S_k \cap E|}{\min(k, |E|)}
\label{eq: hitrate}
\end{align}
In this formula, $|S_k \cap E|$ indicates the number of erroneous sections identified among the top $k$ sections. This metric reflects the ability of PRMs to effectively identify erroneous sections within the top $k$ candidate sections. In Table \ref{tab: hitrate}, the relative performance rankings among different PRMs are quite similar to the results in Table \ref{tab: main}. Additionally, we observe that for $k=3$ and $k=5$, the performance differences between various PRMs are not particularly significant. However, when $k=1$, the Qwen2.5-Math-PRM-7B shows a clear performance advantage. Figure \ref{fig: prm_rank} illustrates the ranking ability of different PRMs for the first incorrect section within the sample, which is generally consistent with the performance evaluation results of HitRate@k.


\begin{tcolorbox}[colback=white!95!gray, colframe=gray!70!black, title=Key Finding]
  HitRate@k evaluation aligns with the main results, with Qwen2.5-Math-PRM-7B demonstrating superior performance in identifying the first incorrect section.
\end{tcolorbox}

\begin{figure}[t]
    \centering
    \includegraphics[width=0.8\textwidth]{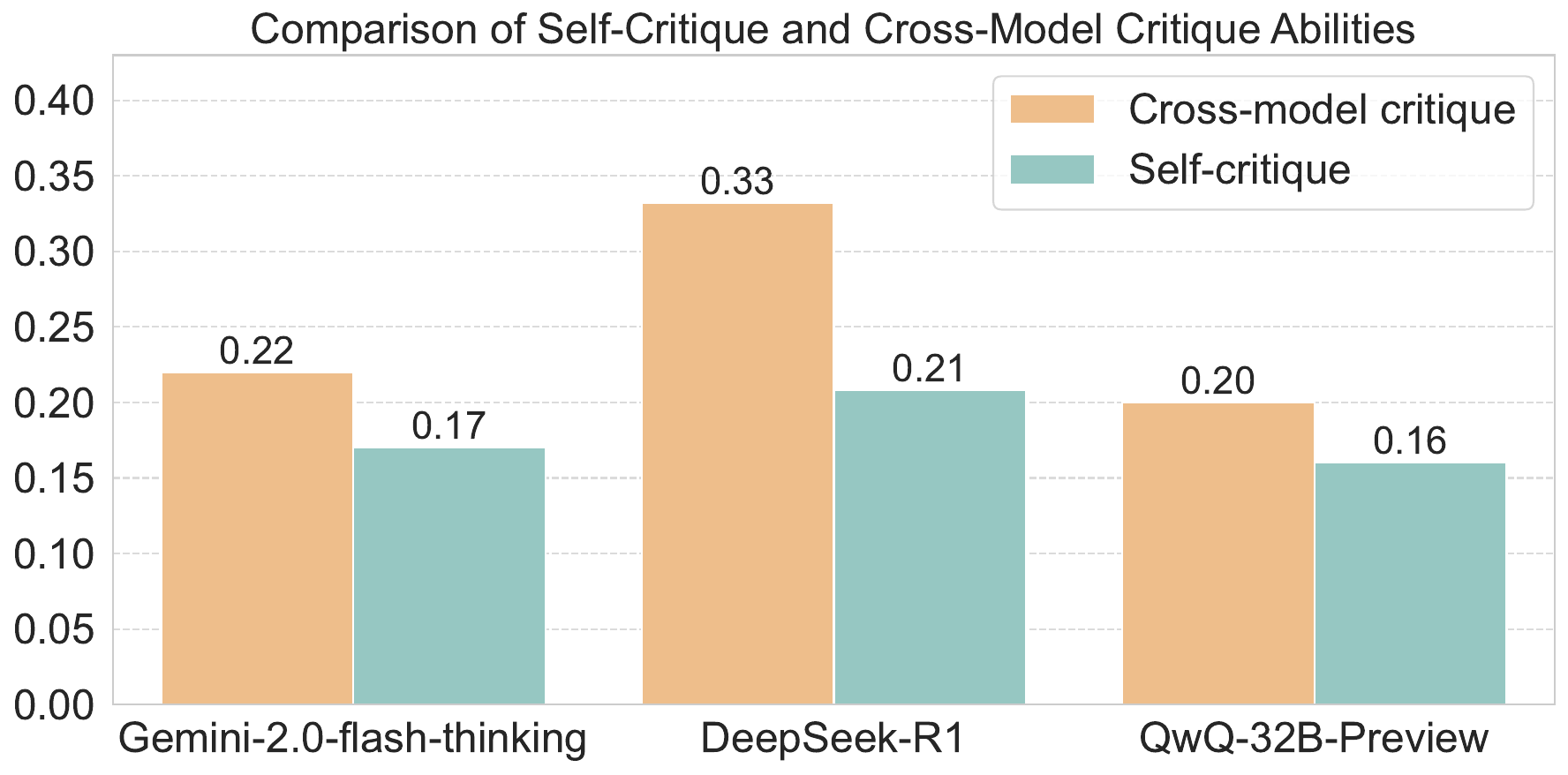}
    \caption{F1-score comparison of self-critique and cross-model critique abilities for different models.}
    \label{fig: self-critic}
\end{figure}

\paragraph{Comparative Analysis of Self-Critique Capabilities of LLMs.} We randomly sample queries based on domains and models that generate the long CoT output, followed by a statistical analysis of the model's performance in evaluating its own outputs as well as those of other models. In Figure \ref{fig: self-critic},  Gemini 2.0 Flash Thinking, DeepSeek-R1, and QwQ-32B-Preview show lower self-critique scores compared to their cross-model critique scores, indicating a prevalent deficiency in self-critic abilities. Notably, DeepSeek-R1 exhibits the largest discrepancy, with a 36\% decrease in self-evaluation compared to evaluations of other models. This suggests models' self-critic abilities remain underdeveloped.

\begin{tcolorbox}[colback=white!95!gray, colframe=gray!70!black, title=Key Finding]
  LLMs demonstrate weaker self-critique performance compared to cross-model critique, highlighting a fundamental limitation in self-critic capabilities.
\end{tcolorbox}

\section{Related Works}

\noindent\textbf{Test-time Scaling.}
Recently, many works have begun to explore the test-time scaling techniques~\citep{openai-o1,snell2024scaling},
can greatly enhance performance by increasing the number of generated tokens. Several  methods~\citep{Guan2025rStarMathSL, chen2024alphamath, hao2023reasoninglanguagemodelplanning, yao2023treethoughtsdeliberateproblem}  have developed the tree search methods to improve reasoning capabilities. 
In addition,
o1-like models (e.g., o1, R1)~\citep{openai-o1, guo2025deepseek, Team2025KimiKS} have investigated the reinforcement learning methods to generate long CoTs and enhance the model performance. Some methods have also begun exploring the intrinsic mechanisms of generating long CoTs~\citep{Yeo2025DemystifyingLC, Wu2024ACS}.


\noindent\textbf{Process Reward Modeling.} 
PRMs demonstrate a significant advantage over traditional outcome-level reward models (ORMs) in enhancing the accuracy of process reasoning~\citep{Lightman2023LetsVS}. The development of an increasing number of PRMs~\citep{Zhang2025TheLO, Xia2024EvaluatingMR, skyworkopeno12024} offers valuable contributions to multi-step reasoning. In addition, the introduction of numerous human-annotated process-level datasets~\citep{cobbe2021gsm8k, Wu2023FineGrainedHF, Li20242DDPOSD} provides essential resources for research. These advancements have collectively spurred exploration in the research direction of generating long CoT.

\noindent\textbf{LLM Critic.}
 The critique capabilities of LLMs have drawn great interests~\citep{Zhang2025CodeCriticBenchAH,liu2025air}. For example, CriticBench \citep{luo2023critique,zheng2024critic} uses LLMs to generate critiques and binary verdicts for input solutions, measuring accuracy by comparing these verdicts to ground truth labels.
 CriticEval \citep{lan2024criticeval} evaluates both feedback and correction qualities.
 Additionally, many works have explored the LLMs' self-critique for improving reasoning~\citep{tyen2023llms,stechly2024self}. 
 For example,
 \citet{huang2023large} have highlighted the challenges in self-correction without external feedback.
 ProCo~\citep{wu2024large} facilitates self-correction and iterative verification processes for better critical abilities. 
\section{Conclusion}
In this paper, we have provided a comprehensive evaluation benchmark called DeltaBench to investigate the limitations of existing o1-like models based on the generated long CoTs and measure the critique qualities of existing LLMs.
Based on DeltaBench, we discuss the specific error analysis of o1-like models and provide a detailed analysis of critic models and PRMs,
where many interesting findings are provided.
Finally, we hope our DeltaBench can not only find the limitations of o1-like models, but also provide guidance to further improve these reasoning models.


\section{Limitations}

While DeltaBench offers a comprehensive evaluation for long CoT reasoning and critique abilities, it has some limitations. 
First, the construction and annotation of the dataset involve high costs, making it challenging to scale to a larger volume of data. Second, although we established a rigorous human annotation mechanism, the process may still introduce subjective biases. Third, as a static benchmark, DeltaBench may not fully capture real-time advancements. Addressing these limitations will be a key focus of our future work.

\bibliographystyle{unsrtnat}
\bibliography{references} 

\clearpage
\onecolumn
\appendix
\onecolumn

\section{Details of Dataset Construction}
\label{app: dataset}

\subsection{Data Sources}
\label{app: data_source}

\begin{table}[ht]
    \centering
    \small
    \begin{tabular}{c|l}
        \toprule
        \textbf{Domain} & \textbf{Source} \\ 
        \midrule
        Math & \begin{tabular}[c]{@{}l@{}} 
                   MATH-500~\citep{Lightman2023LetsVS}, OlympiadBench~\citep{He2024OlympiadBenchAC},\\ Omni-MATH~\citep{Gao2024OmniMATHAU}, AIME, AMC23, CollegeMath
               \end{tabular} \\ 
               \midrule
        Programming & \begin{tabular}[c]{@{}l@{}} 
                          CodeForce, BigCodeBench~\citep{zhuo2024bigcodebench}, LiveCodeBench~\citep{Jain2024LiveCodeBenchHA}, \\ 
                          FullStackBench~\citep{Cheng2024FullStackBE}, NaturalCodeBench~\citep{Zhang2024NaturalCodeBenchEC},\\ USACO~\citep{Shi2024CanLM} 
                      \end{tabular} \\
                      \midrule
        PCB & \begin{tabular}[c]{@{}l@{}} 
                  GPQA~\citep{Rein2023GPQAAG}, OlympiadBench~\citep{He2024OlympiadBenchAC}, \\ AGIEval~\citep{Zhong2023AGIEvalAH}
              \end{tabular} \\ 
              \midrule
        General Reason & \begin{tabular}[c]{@{}l@{}} 
                             ZebraLogicBench~\citep{Lin2025ZebraLogicOT}, KOR-Bench~\citep{Ma2024KORBenchBL}, \\
                             Geeksforgeeks Puzzles, CS Interview Questions, China Civil Service Exam Questions 
                         \end{tabular} \\ 
                         \bottomrule
    \end{tabular}
    \vspace{+3mm}
    \caption{Query-related data source statistics. Parts without citations come from open internet resources.}
    \label{tab: query_source}
\end{table}
Table \ref{tab: query_source} shows the original data sources from which we extracted high-quality queries and their corresponding solutions, along with additional information (for code data, test cases were extracted).


\subsection{Details on Dataset Annotation}
\label{app:anno}
\label{app: profile}

The average cost for annotating each data unit was approximately \textit{\$15}. 
The annotation process is organized into three phases, each with specific goals and criteria:

\paragraph{Initial Assessment.} 
Annotators initially verify the quality of the question and subsequently evaluate the correctness of the model’s final response.

\paragraph{Section-Level Evaluation.}
The long CoT is divided into sections, each corresponding to a specific sub-task, such as problem analysis, verification of calculation results, and summarization. This phase requires the annotator to check and annotate each section individually.
The annotation process and examples are shown in Figure \ref{fig: label_case}



\paragraph{Quality Assurance and Validation.}
We have established a strict quality control process to ensure the high quality and consistency of annotations. Each data is assigned three initial annotators, two junior reviewers, and an additional five people who are responsible for overall spot checks. Outsourced personnel and external contractors responsible for annotations receive unified training. We regularly check the consistency and quality between annotators and repeatedly discuss and improve annotation protocols during the annotation process to make the standards more perfect. This process ensures the generation of high-quality annotations and minimizes subjective bias.

\paragraph{Profile of Annotation Persons.}

In this dataset annotation project, we engaged a diverse group of annotators through three distinct sources. We employed a set of external contractors directly recruited by us and collaborated with two additional suppliers to provide annotation services. The dataset was divided into three parts, with certain sections deliberately overlapping to facilitate cross-validation. This overlap allows us to compute the annotation consistency rate, and if the results do not meet the required standards, revisions are necessitated.

Our annotator pool is composed of highly qualified individuals: 23 Master's degree holders and 6 Ph.D. holders in Mathematics, 18 Master's graduates in Computer Science, 7 Master's and 2 Ph.D. graduates in Physics, 7 Master's and 3 Ph.D. graduates in Chemistry, and 6 Master's and 2 Ph.D. degree holders in Biology. We employ a rotational system to randomly assign two individuals from each academic field to serve as reviewers. Additionally, 5 algorithm specialists are tasked with conducting spot checks on the annotations. This meticulous selection and review process ensures superior data quality and reliability for subsequent analyses.

\subsection{Details on Assessment}
\label{app:asses}

\paragraph{Math} For mathematical queries, we employ a combination of rules and LLMs to evaluate the correctness of the provided solutions. Rule-based systems verify the validity of numerical calculations, while LLMs ensure that reasoning steps adhere to established mathematical principles. This dual approach guarantees high accuracy in error detection within the solutions.
\paragraph{Programming} For programming tasks, we utilize sandbox testing environments alongside LLM-based evaluations.  Specifically, we utilize SandboxFusion~\citep{Cheng2024FullStackBE} as our testing environment. The solution is initially executed in the sandbox environment. Subsequently, the test case, the sandbox environment's feedback output, and the code are provided to the LLM to determine the correctness of the answer.
\paragraph{PCB} Due to the straightforward nature of answers in these domains, we exclusively rely on LLM judgments, which offer high accuracy in assessing correctness.
\paragraph{General Reasoning} Similarly, for general reasoning questions, LLM judgments are employed to effectively and accurately assess solution validity.

\subsection{Findings in Data Preprocessing}
\label{app: data_preprocess}
During the data preprocessing stage, we identified several issues with the data collected from open-source datasets. These issues included incomplete queries, incorrect solutions, and excessively high query similarity. To address these problems, we applied a combination of manual review and LLM (Large Language Model) validation to filter out low-quality data. Additionally, for code data specifically, we observed that different sources and types of data sometimes included test cases, while others did not, and the formats of these test cases were inconsistent. To tackle these inconsistencies, we used GPT-4 to filter the data for quality and to extract test cases, standardizing them into executable code for SandboxFusion. This allowed us to conduct uniform sandbox verification to ensure data accuracy.

\subsection{Sections Division}
\label{app: section_div}

\begin{figure}[!htbp]
    \centering
    \begin{subfigure}{0.48\textwidth}
        \centering
        \includegraphics[width=0.9\textwidth]{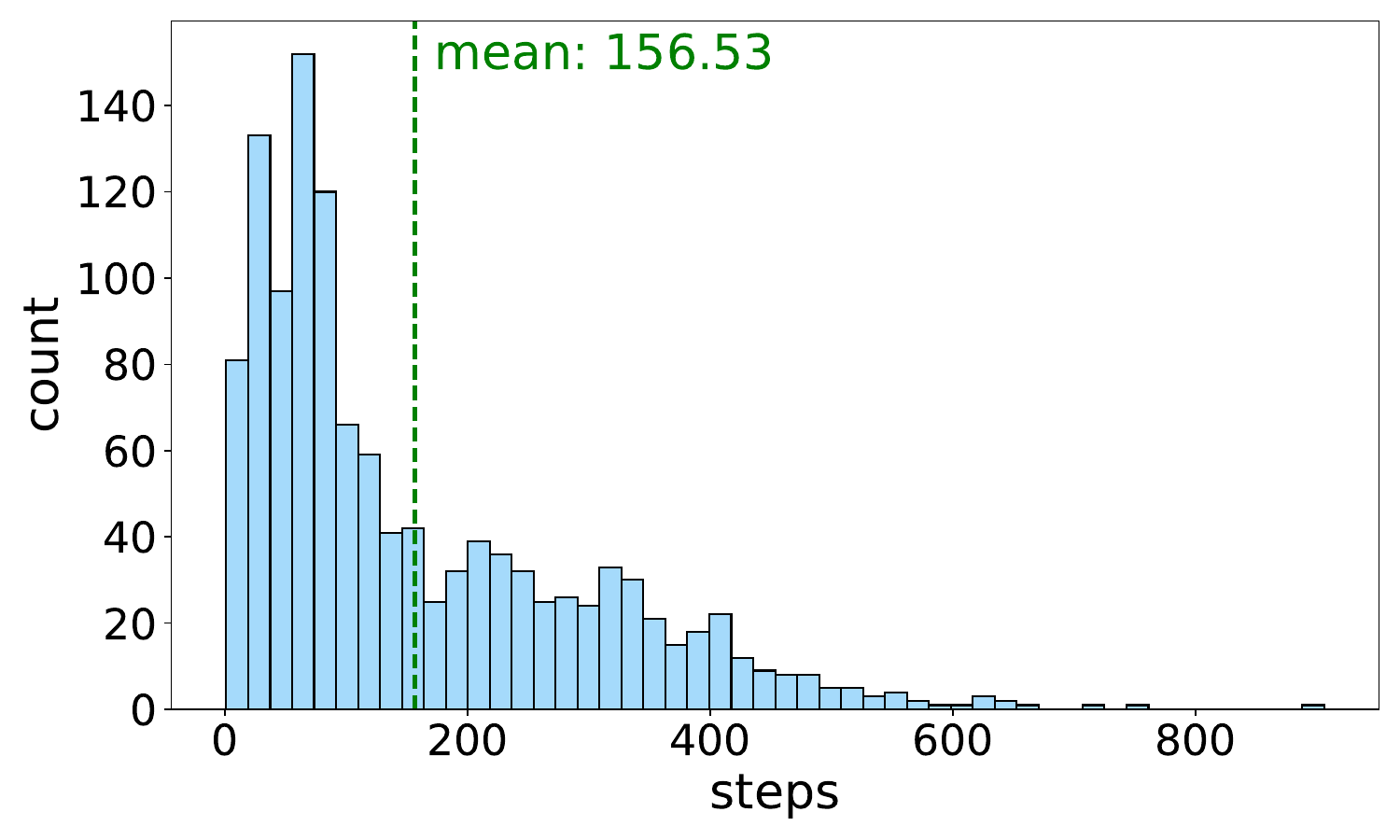}
        \caption{Distribution of the number of steps in long CoT(divided by "\texttt{\textbackslash n\textbackslash n}").}
        
        \label{fig:longcot_origin_steps_dist}
    \end{subfigure}
    \hfill
    \begin{subfigure}{0.48\textwidth}
        \centering
        \includegraphics[width=0.9\textwidth]{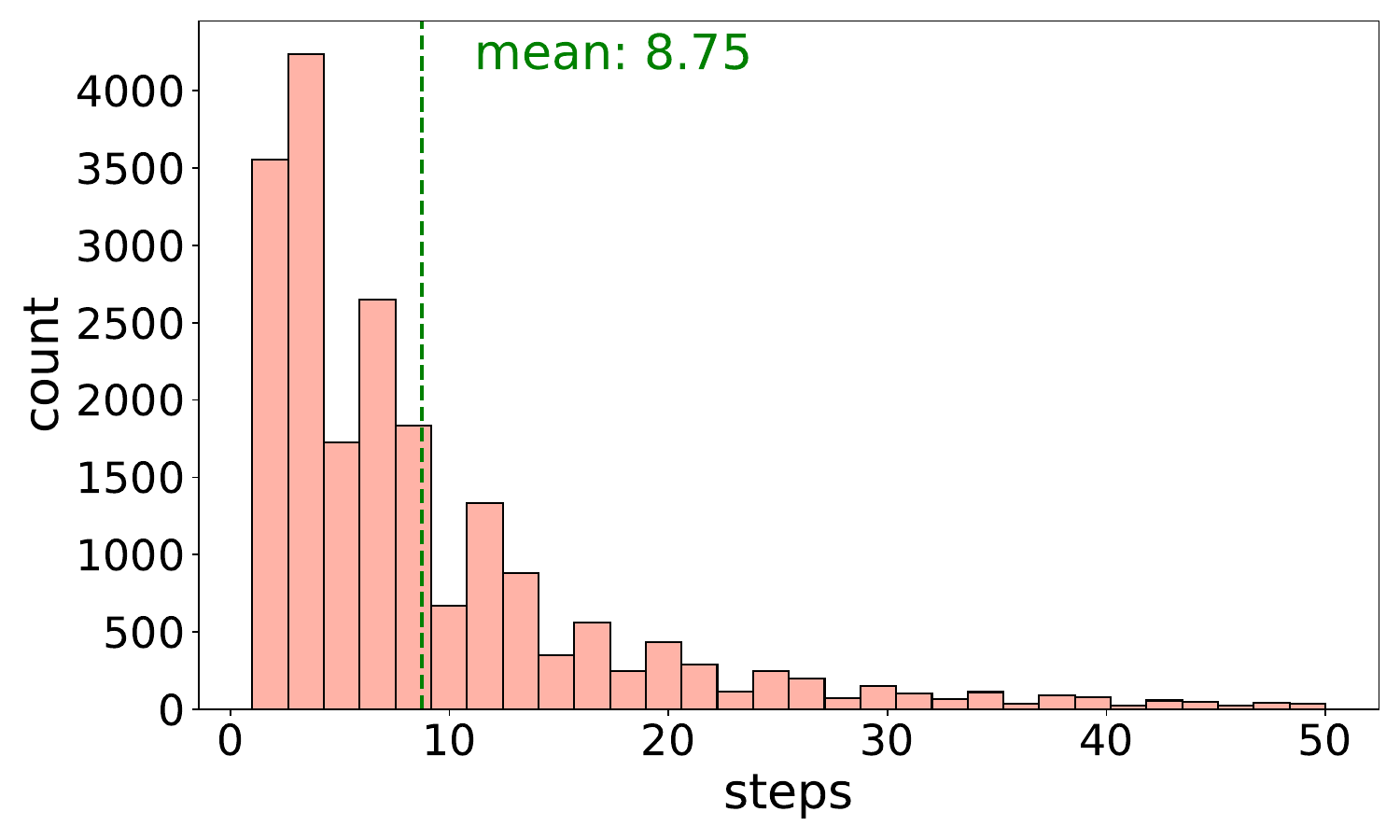}
        \caption{Distribution of the number of steps contained in each divided section.}
        \label{fig:section_step_dist}
    \end{subfigure}
    \caption{Statistical distribution of steps in long CoT.}
    \label{fig:longcot_steps_dist}
\end{figure}

Figure \ref{fig: section_div} illustrates the prompt for dividing sections along with examples of the resulting divisions. Several steps involved in addressing an atomic problem or exploring an idea are grouped into the same section. The specific outcome of the division is influenced by various factors, such as the task domain. However, compared to a purely long CoT, this approach is more user-friendly for human annotation.

Furthermore, to prevent sections from becoming overloaded with too many steps, which would increase the complexity of the annotation process, we iteratively divide sections that exceed $50$ steps. Figure \ref{fig:longcot_steps_dist} displays the distribution of steps in the original long CoT (subfigure \ref{fig:longcot_origin_steps_dist}) and the distribution of steps in each divided section (subfigure \ref{fig:section_step_dist}). Before sectioning, annotators are required to review each individual step, which can be exceedingly challenging for long CoTs with numerous steps. By dividing the sections, annotators can proceed on a section-by-section basis, making the process more comprehensible and significantly reducing the difficulty of annotation.

\subsection{Validation of Long CoT Correctness}
\label{app: correct_assess}

\begin{table}[!ht]
    \centering
    \begin{tabular}{c|ccc}
    \hline
        Domain & Filtered Total Num & Correct Num & Accuracy(\%) \\ \hline
        Math & 7534 & 5413 & 71.84 \\ 
        Programming & 2103 & 1276 & 60.67 \\ 
        PCB & 4517 & 2626 & 58.13 \\
        General Reasoning & 1981 & 1160 & 58.56 \\ \hline
    \end{tabular}
    \vspace{+3mm}
    \caption{Accuracy statistics for generated long CoT responses for filtered high-quality queries.}
    \label{tab: longcot_acc}
\end{table}

For the filtered high-quality queries, we use a mix of various o1-like models, including QwQ-32B-Preview, DeepSeek-R1, and  Gemini 2.0 Flash Thinking, to generate the corresponding long CoT. We then use LLM-as-a-judge and a sandbox testing environment to validate the accuracy of the long CoT generated by these o1-like models, obtaining the native erroneous long CoT for subsequent human annotation.

Table \ref{tab: longcot_acc} shows the accuracy of the generated long CoT. It can be seen that o1-like models enhanced with reinforcement learning in math and programming perform slightly better in these two areas compared to general reasoning and PCB.

\subsection{Statistics on Category Distribution}
\label{app: category_distribution}

\begin{table}[!ht]
    \centering 
    \begin{tabular}{llc}
        \toprule
        \textbf{Domain} & \textbf{Subcategory} & \textbf{Number} \\ 
        \midrule
        \multirow{7}{*}{Math} 
        & Discrete Mathematics & 144 \\ 
        & Number Theory & 104 \\ 
        & Geometry & 101 \\ 
        & Others & 74 \\ 
        & Calculus and Analysis & 58 \\ 
        & Statistics and Other Decision Science & 45 \\ 
        & Algebra & 36 \\
        \midrule
        \multirow{7}{*}{Programming} 
        & Basic Programming & 133 \\ 
        & Mathematics & 86 \\ 
        & Advanced Programming & 48 \\ 
        & Data Analysis & 41 \\ 
        & Desktop and Web Development & 27 \\ 
        & Others & 24 \\ 
        & Software Engineering & 14 \\ 
        \midrule
        \multirow{3}{*}{PCB} 
        & Chemistry & 64 \\ 
        & Physics & 63 \\ 
        & Biology & 27 \\ 
        \midrule
        \multirow{7}{*}{General Reasoning} 
        & Logical Reasoning & 56 \\ 
        & Symbolic Reasoning & 28 \\ 
        & Quantitative Reasoning & 24 \\ 
        & Strategic Reasoning & 12 \\ 
        & Common Sense Reasoning & 9 \\ 
        & Spatio-temporal Reasoning & 9 \\ 
        & Others & 9 \\
        \midrule
        & Total & 1236 \\
        \bottomrule
    \end{tabular}
    \vspace{+3mm}
    \caption{Detailed categories of DeltaBench and corresponding data volume statistics.}
    \label{tab: data_category}
\end{table}

Table \ref{tab: data_category} shows the subcategories and corresponding data volumes of DeltaBench across various domains. In obtaining queries and annotations, we strive to ensure balance across categories while also balancing annotation difficulty and accuracy.





\subsection{Analysis of Other Evaluation Metrics}
\label{app: other_evaluation}

\FloatBarrier
\begin{table}[!h]
\centering
\resizebox{0.8\textwidth}{!}{ 
\begin{tabular}{cccc}
\toprule
Model & F1-Score & First Error Acc. & Any Error Acc. \\
\midrule
GPT-4-turbo-128k & 40.76 & 57.04 & 69.17 \\
GPT-4o & 30.85 & 36.89 & 50.89 \\
DeepSeek-V3 & 27.33 & 31.72 & 42.39 \\
Qwen2.5-32B-Instruct & 26.73 & 30.58 & 42.23 \\
DeepSeek-R1 & 28.43 & 29.94 & 40.78 \\
Qwen2.5-7B-Instruct & 18.63 & 22.25 & 30.74 \\
GPT-3.5 & 7.98 & 6.15 & 11.65 \\
\bottomrule
\end{tabular}
}
 \vspace{+3mm}
\caption{The table compares different accuracy metrics for each model. 'First Error Acc.' is the accuracy in identifying the first error, and 'Any Error Acc.' is the accuracy in detecting any error.}
\label{tab:accuracy_comparison}
\end{table}
\FloatBarrier

In Table ~\ref{tab:accuracy_comparison}, we present the performance of several models across different accuracy metrics: F1-Score, First Error Accuracy, and Any Error Accuracy. These metrics evaluate the models' ability to identify the first error and detect any error within a given sequence. A key observation is that the relative rankings of the models across the First Error Accuracy and Any Error Accuracy metrics closely align with their F1-Score. This consistency across different evaluation measures highlights the robustness of the F1-Score as a comprehensive indicator of model performance and suggests a strong correlation between the ability to detect the first error and the ability to identify any error in the sequence. Additionally, GPT-4-turbo consistently outperforms the other models, regardless of the evaluation metric used. Its Any Error Accuracy reaches 69\%, significantly higher than the other models in the comparison. This finding underscores the model’s superior performance in error recognition, yet it also points to the limitations that remain in current LLMs.

\begin{table}[!ht]
    \centering
    \small
    \begin{tabular}{cccccc}
    \bottomrule
        Model & Quantile & Threshold & prec & recall & F1 \\
        \midrule
        Qwen/Qwen2.5-Math-PRM-7B & 5\% & 0.2168 & 39.81 & 73.86 & 46.48 \\ 
        Qwen/Qwen2.5-Math-PRM-72B & 5\% & 0.2119 & 33.51 & 65.11 & 40.44 \\ 
        RLHFlow/Llama3.1-8B-PRM-Deepseek-Data & 5\% & 0.2021 & 24.19 & 56.1 & 30.88 \\ 
        RLHFlow/Llama3.1-8B-PRM-Mistral-Data & 5\% & 0.2949 & 23.18 & 51.46 & 29.68 \\ 
        Skywork/Skywork-o1-Open-PRM-Qwen-2.5-1.5B & 5\% & 0.0303 & 19.48 & 46.76 & 24.45 \\ 
        Skywork/Skywork-o1-Open-PRM-Qwen-2.5-7B & 5\% & 0.0278 & 18.68 & 46.14 & 23.46 \\ 
        \bottomrule
    \end{tabular}
     \vspace{+3mm}
    \caption{Performance of PRMs using the overall reward quantile as the threshold.}
\label{tab:prm_quantile}
\end{table}

For PRMs, aside from outlier detection, we also experimented with evaluating using a fixed threshold based on quantiles. Specifically, we used the ascending 5\% quantile of all rewards on DeltaBench as the threshold, considering sections below this value as incorrect. The evaluation results are shown in table \ref{tab:prm_quantile}. However, compared to outlier detection, we found that this approach overestimates the performance of PRMs. This is because using a quantile as a threshold effectively forces PRMs to consider a fixed proportion of sections as incorrect.


\section{Error Classification}
\label{app: error_classification}

\begin{figure*}[t]
\centering
\vspace{-10mm}
\includegraphics[width=0.95\linewidth]{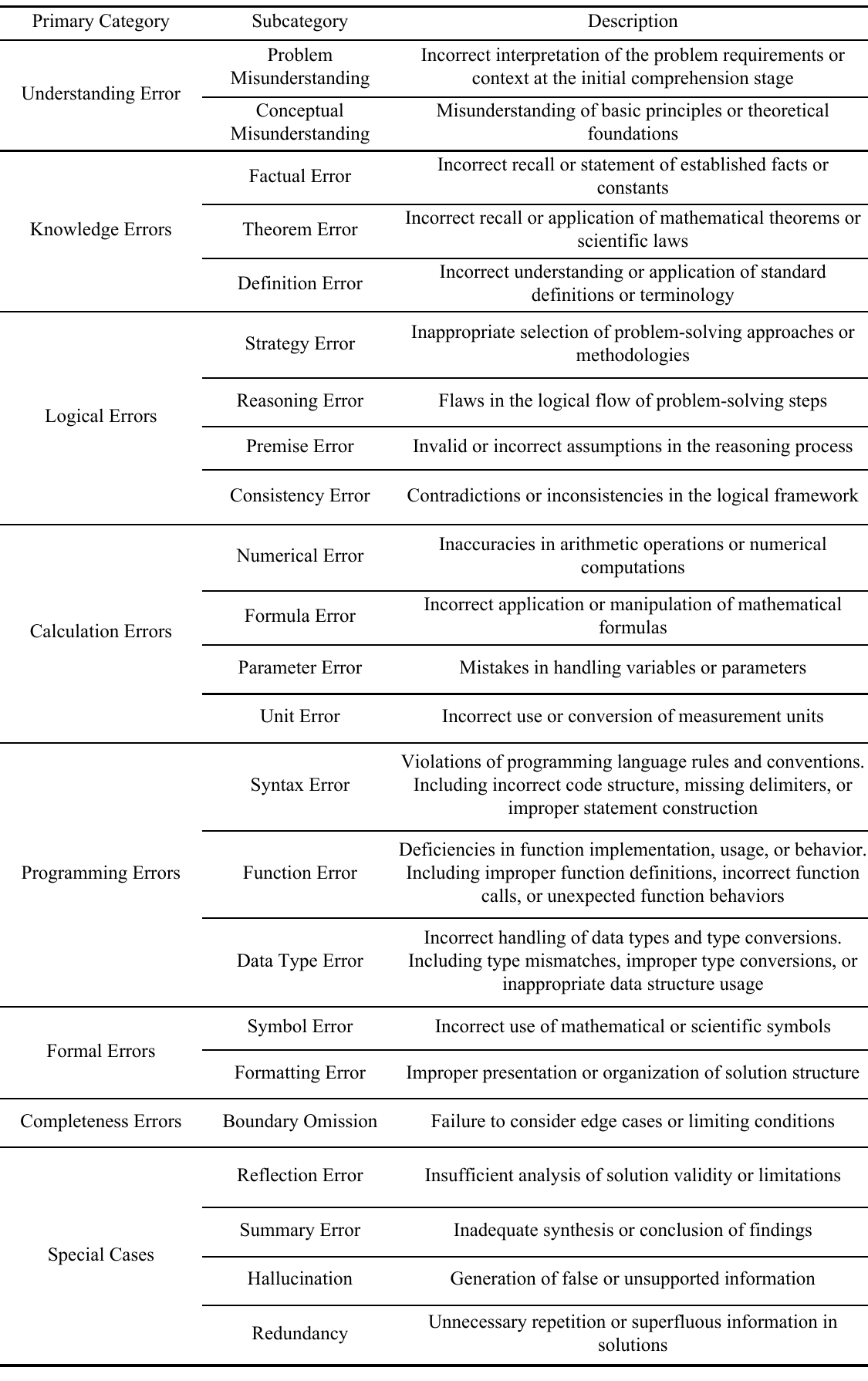}
\caption{Categories of Errors in o1-like Models.}
\label{fig: error_category}
\vspace{-4mm}
\end{figure*}

In Figure \ref{fig: error_category}, we conclude a detailed error classification based on human annotations of the errors contained in the model’s answers.

\section{Analysis of Underperforming Model}
\label{app: underperforming}

\FloatBarrier
\begin{figure*}[!htbp]
\centering
\vspace{-10mm}
\includegraphics[width=0.6\linewidth]{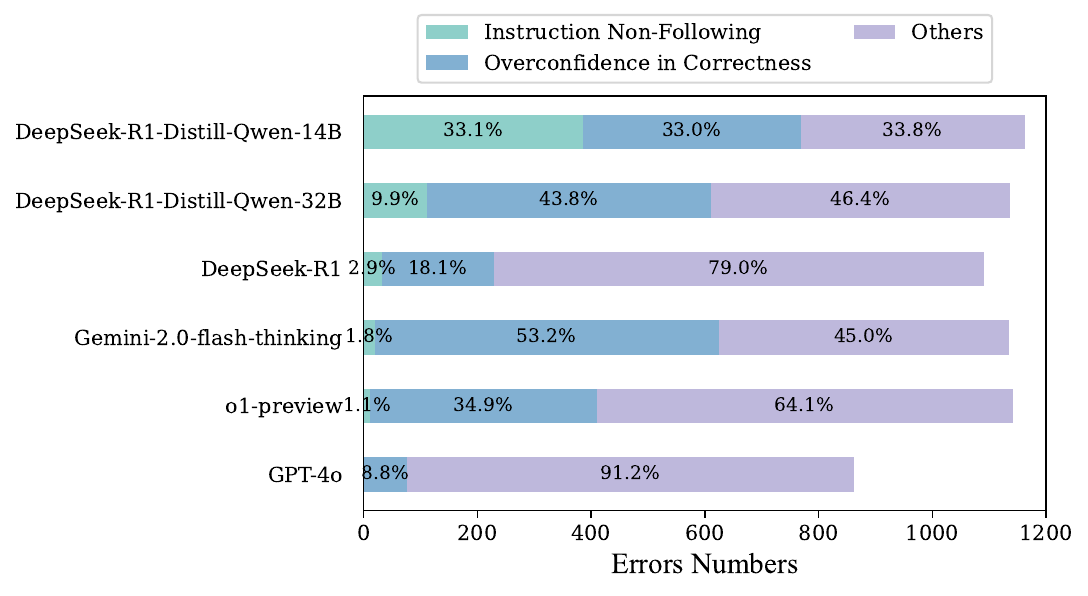}
\caption{Distribution of error causes for each model.}
\label{fig: wrong_types}
\end{figure*}
\FloatBarrier

We classify the errors produced by underperforming models into three categories: (1) Not following instructions, where models fail to follow the critique instructions; (2) Overconfidence in Correctness, where models incorrectly assume the sample contains no errors; (3) Other errors, such as false negatives, where models incorrectly flag accurate reasoning as erroneous, and error misidentification, where models inaccurately determine the type or location of errors.

As illustrated in Figure ~\ref{fig: wrong_types}, the following observations can be made: (1) DeepSeek-R1-Distill series exhibit significant instruction-following deficiencies. These models tend to directly answer questions rather than critically evaluate the correctness of responses, indicating a potential overfitting problem. (2) o1-like models like DeepSeek-R1 and o1-preview-0912 also demonstrate notable instruction-following challenges, albeit to a lesser extent than GPT-4o. (3) A notable trend among the o1-preview and  Gemini 2.0 Flash Thinking models was their tendency to assume that samples were correct without thorough evaluation. This overconfidence in correctness was a significant contributor to their error rates. In contrast, GPT-4o demonstrates superior performance, with no instances of "Instructions not followed" errors and a relatively low number of "Overconfidence in Correctness" errors.
\label{sec:appendix}
\end{document}